\documentclass[10pt,twocolumn,letterpaper]{article}

\usepackage{cvpr}              

\usepackage{url}            
\usepackage{booktabs}       
\usepackage{multirow}
\usepackage{bbding} 

\usepackage{amsfonts}       
\usepackage{amsthm,amsmath,amssymb}
\usepackage{bbm}

\usepackage{mathrsfs}
\usepackage{nicefrac}       
\usepackage{microtype}      
\usepackage{xcolor}         

\usepackage{graphicx}
\usepackage{float}
\usepackage{subfloat}
\usepackage{bbding}
\usepackage{lipsum}

\def\eg{\emph{e.g.~}}
\def\ie{\emph{i.e.~}}

\usepackage{enumitem}

\usepackage[pagebackref,breaklinks,colorlinks]{hyperref}

\usepackage[capitalize]{cleveref}
\crefname{section}{Sec.}{Secs.}
\Crefname{section}{Section}{Sections}
\Crefname{table}{Table}{Tables}
\crefname{table}{Tab.}{Tabs.}

\def\cvprPaperID{4630} 
\def\confName{CVPR}
\def\confYear{2022}

\newenvironment{packed_itemize}{
	\begin{itemize}[leftmargin=*]
		\setlength{\itemsep}{-0.1pt}
		\setlength{\parskip}{-0.1pt}
		\setlength{\parsep}{-1pt}
	}{\end{itemize}} 

\newcommand\Mark[1]{\textsuperscript#1}

\begin{document}

\title{Delving Deep into the Generalization of Vision Transformers \\ under Distribution Shifts}

\author{Chongzhi Zhang\Mark{1}\Mark{,}\Mark{*}, Mingyuan Zhang\Mark{2}\Mark{,}\Mark{*}, Shanghang Zhang\Mark{3}\Mark{,}\Mark{*}, Daisheng Jin\Mark{1}, Qiang Zhou\Mark{4},\\Zhongang Cai\Mark{2}\Mark{,}\Mark{5}, Haiyu Zhao\Mark{2}\Mark{,}\Mark{5}, Xianglong Liu\Mark{1}, Ziwei Liu\Mark{2}\Mark{,}\Mark{\Envelope}
\\
\Mark{1}Beihang University\quad\Mark{2}S-Lab, Nanyang Technological University
\\\Mark{3}Peking University\quad\Mark{4}AIR, Tsinghua University\quad\Mark{5}Shanghai AI Laboratory
}

\maketitle
\newcommand\blfootnote[1]{%
\begingroup
\renewcommand\thefootnote{}\footnote{#1}%
\addtocounter{footnote}{-1}%
\endgroup
}

\blfootnote{\Mark{*}These authors contributed equally to this work.}
\blfootnote{\Mark{\Envelope}Corresponding author.}


\vspace{-10pt}
\begin{abstract}
  Vision Transformers (ViTs) have achieved impressive performance on various vision tasks, yet their generalization under distribution shifts (DS) is rarely understood. In this work, we comprehensively study the out-of-distribution (OOD) generalization of ViTs. 
  For systematic investigation, we first present a taxonomy of DS.
  We then perform extensive evaluations of ViT variants under different DS and compare their generalization with Convolutional Neural Network (CNN) models. 
  Important observations are obtained: \textbf{1)} ViTs learn weaker biases on backgrounds and textures, while they are equipped with stronger inductive biases towards shapes and structures, which is more consistent with human cognitive traits. Therefore, ViTs generalize better than CNNs under DS. With the same or less amount of parameters, ViTs are ahead of corresponding CNNs by more than 5\% in top-1 accuracy under most types of DS. \textbf{2)} As the model scale increases, ViTs strengthen these biases and thus gradually narrow the in-distribution and OOD performance gap. 
  To further improve the generalization of ViTs, we design the Generalization-Enhanced ViTs (GE-ViTs) from the perspectives of adversarial learning, information theory, and self-supervised learning. By comprehensively investigating these GE-ViTs and comparing with their corresponding CNN models, 
  we observe: \textbf{1)} For the enhanced model, larger ViTs still benefit more for the OOD generalization. \textbf{2)} GE-ViTs are more sensitive to the hyper-parameters than their corresponding CNN models. 
  We design a smoother learning strategy to achieve a stable training process and obtain performance improvements on OOD data by 4\% from vanilla ViTs.
  We hope our comprehensive study could shed light on the design of more generalizable learning architectures. Codes are available \href{https://github.com/Phoenix1153/ViT_OOD_generalization}{here}.
  
  
\end{abstract}

\section{Introduction}


Recently, transformer has made remarkable achievements in vision tasks, such as
\eg image classification \cite{chen2020generative,dosovitskiy2020image,touvron2020training}, object detection \cite{DBLP:conf/eccv/CarionMSUKZ20,zhu2021deformable}, and image processing \cite{chen2020pre}. 
Despite the encouraging performance achieved on standard benchmarks and several properties revealed in recent works \cite{paul2021vision,bai2021transformers,caron2021emerging,naseer2021intriguing}, the generalization ability of Vision Transformers (ViTs) is still less understood.
While the traditional train-test scenario assumes the test data for model evaluation are independent identically distributed (IID) with sampled training data, this assumption does not always hold in real-world scenarios. 
Thus, out-of-distribution (OOD) generalization is a highly desirable capability of machine learning models.
Recent works indicate current CNN architectures generalize poorly on various distribution shifts (DS) \cite{geirhos2018imagenettrained,hendrycks2018benchmarking,hendrycks2020many}, whereas the investigation on ViTs remains scarce. Therefore, in this paper, we mainly focus on delving deep into the OOD generalization of ViTs under DS.

To comprehensively study the OOD generalization ability of ViTs, we first define a categorization of commonly appearing DS based on the modified semantic concepts in images. Generally, an image for classification contains a foreground object and background information. The foreground object consists of hierarchical semantic concepts including pixel-level elements, object textures, and shapes, object parts, and object itself \cite{zeiler2014visualizing}. A distribution shift usually causes variance on one or more semantics and we thus present a taxonomy of DS into four conceptual groups: background shifts, corruption shifts, texture shifts, and style shifts. 


With the taxonomy of DS, we investigate the OOD generalization of ViTs by comparison with CNNs in each case. 
While models are desired to generalize to arbitrary OOD scenarios, the no-free-lunch theorem for machine learning \cite{wolpert1995no,baxter2000model,goyal2020inductive} demonstrates that there is no entirely general-purpose learning algorithm, and that any learning algorithm implicitly or explicitly will generalize better on some distributions and worse on others. Thus some set of inductive biases are demanded to acquire generalization. 
Hence, to achieve human-level generalization capability, machine learning models are supposed to have inductive biases that are most relevant to the human prior in the world. 
There have been many attempts to inject inductive biases into deep learning models that humans may exploit for the cognition operating at the level of conscious processing, \eg the convolution \cite{lecun1995convolutional} and self-attention mechanism \cite{DBLP:conf/nips/VaswaniSPUJGKP17}. 
Therefore, we examine whether transformers are equipped with inductive biases that are more related to human cognitive traits to better investigate the generalization properties of ViTs under DS.
Extensive evaluations reveal the following observations on the OOD generalizations of ViTs: \textbf{1)} ViTs learn weaker biases on backgrounds and textures, while they are equipped with stronger inductive biases towards shapes and structures, which is more consistent with human cognitive traits. Therefore, ViTs generalize better than CNNs in most cases. Specifically, ViT not only achieves better performance on OOD data but also has smaller generalization gaps between IID and OOD datasets. \textbf{2)} As the model scale increases, ViTs strengthen these biases and thus gradually narrow the IID and OOD generalization gaps, especially in the case of corruption shifts and background shifts. In other words, larger ViTs are better at diminishing the effect of local changes. \textbf{3)} ViTs trained with larger patch size deal with texture shifts better, yet are inferior in other cases.

After validating the superiority of ViTs in dealing with OOD data, we focus on further improving their generalization capacity. Specifically, we design Generalization-Enhanced ViTs (GE-ViTs) from the perspectives of adversarial training \cite{ganin2015unsupervised}, information theory \cite{saito2019semi} and self-supervised learning \cite{yue2021prototypical}. 
Equipped with GE-ViTs, we achieve significant performance boosts towards OOD data by 4\% from vanilla ViTs. By performing an in-depth investigation on different models, we draw the following conclusions: \textbf{1)} For the enhanced transformer models, larger ViTs still benefit more for the OOD generalization. \textbf{2)} GE-ViTs are more sensitive to the hyper-parameters than their corresponding CNN models.  

\section{Related Work}




\noindent\textbf{Vision Transformers.}
Recently, Transformers have been applied to various vision tasks including image classification \cite{chen2020generative,dosovitskiy2020image,touvron2020training}, object detection \cite{DBLP:conf/eccv/CarionMSUKZ20,zhu2021deformable}, segmentation \cite{wang2020end} and image processing \cite{chen2020pre}. Among them, the Vision Transformer (ViT) \cite{dosovitskiy2020image} is the first fully-transformer model applied for image classification and competitive with state-of-the-art CNNs. It heavily relies on large-scale datasets for model pre-training, requiring huge computation resources. 
Later, \cite{touvron2020training} propose the Data-efficient image Transformer (DeiT), which achieves competitive results against the state-of-the-art CNNs on ImageNet without external data by simply changing training strategies from ViT. Due to its efficiency, we use this family of models to investigate generalizations of Vision Transformers in this paper.

\noindent\textbf{Out-of-distribution Generalization.}
Attracting much attention recently, various works have been proposed for OOD generalization under different settings. Most domain adaptation literatures aim at promoting the model's performance under distribution shift with access to the unlabeled target data~\cite{ganin2015unsupervised,long2015learning,sun2019unsupervised}. 
Another setting for OOD generalization concentrates on learning representations without access to target data, commonly referred as domain generalization \cite{li2018domain,volpi2018generalizing,dou2019domain}. In addition, some recent works model OOD generalization on their newly-built benchmarks~\cite{hendrycks2018benchmarking,geirhos2018imagenettrained,hendrycks2020many}. 
Though recent works~\cite{paul2021vision,bai2021transformers,caron2021emerging,naseer2021intriguing} have studied several properties of ViTs, the generalization of ViTs is still under explored.

\section{Distribution Shifts and Evaluation Protocols}



\subsection{Taxonomy of Distribution Shifts} \label{sec:categorize}


To make an extensive study on OOD generalization, we build the taxonomy of DS upon what kinds of semantic concepts are modified from the original image. 
Therefore, we divide the DS into four cases: background shifts, corruption shifts, texture shifts and style shifts, as shown in \cref{table:Shift Taxonomy}. The elaborately divided DS permit us to investigate model biases towards every visual cue respectively.


\begin{table}
  \caption{\textbf{Illustration of our taxonomy of DS.} We build the taxonomy upon what kinds of semantic concepts are modified from the original image and divide the DS into four cases: background shifts, corruption shifts, texture shifts, and style shifts. \textcolor{red}{\checkmark} denotes the unmodified vision cues under certain type of DS.}
  \vspace{-5pt}
  \label{table:Shift Taxonomy}
  \centering
  \resizebox{1.0\linewidth}{!}{
  \begin{tabular}{cccccc}
    \toprule
    \multirow{2}{*}{Shift Type} & \multirow{2}{*}{background} & \multicolumn{4}{c}{foreground} \\ \cline{3-6}
    ~ & ~ & pixel & texture & shape & structure \\ 
    \midrule
    Background Shift &   & \textcolor{red}{\checkmark}  & \textcolor{red}{\checkmark} & \textcolor{red}{\checkmark} & \textcolor{red}{\checkmark} \\
    Corruption Shift &   &   & \textcolor{red}{\checkmark} & \textcolor{red}{\checkmark} & \textcolor{red}{\checkmark} \\
    Texture Shift &   &   &  & \textcolor{red}{\checkmark} & \textcolor{red}{\checkmark} \\
    Style Shift &   &   &  &  & \textcolor{red}{\checkmark} \\
    \bottomrule
  \end{tabular}
  }
  \vspace{-10pt}
\end{table}

\begin{packed_itemize}

\item \textbf{Background Shifts.} Image backgrounds are usually regarded as auxiliary cues in assigning images to corresponding labels in the image classification task. However, previous works have demonstrated that backgrounds may dominate in prediction \cite{rosenfeld2018elephant,barbu2019objectnet}, which is undesirable to us. We focus on the model's invariance towards background change and thus define the background shifts. \emph{ImageNet-9} \cite{xiao2020noise} is adopted for background shifts. 

\item \textbf{Corruption Shifts.} The concept of corruption was proposed in \cite{hendrycks2018benchmarking}, which stands for those naturally occurring vicinal impurities mixed in images. These corruptions either come from environmental influence during the shooting stage or from the image processing stage. We define these cases as corruption shifts, which only impact on object pixel-level elements while can still cause models obvious performance decrease. \emph{ImageNet-C} \cite{hendrycks2018benchmarking} is used to examine generalization ability under corruption shifts. 

\item \textbf{Texture Shifts.} Generally, the texture gives us information about the spatial arrangement of the colors or intensities in an image, which is critical for the classifiers in obtaining a correct prediction. Thus, a replacement of object textures can influence model prediction. We define these variations as texture shifts. \emph{Cue Conflict Stimuli} and \emph{Stylized-ImageNet} \cite{geirhos2018imagenettrained} are used to investigate generalization under texture shifts. 

\item \textbf{Style Shifts.} Typically, style is a complicated concept determined by the characteristics that describe the artwork, such as the form, color, composition, etc. The variance of style often reflects in multiple concept levels, including texture, shape and object part, etc. 
\emph{ImageNet-R} \cite{hendrycks2020many} and \emph{DomainNet} \cite{peng2019moment} are used for the case of style shifts.


\end{packed_itemize}

\subsection{Model Zoo}

\begin{packed_itemize}

\item \textbf{Vision Transformer.} We follow the implementation in DeiT \cite{touvron2020training} and choose a range of models with different scales for experiments. The ViT architecture takes as input a grid of non-overlapping contiguous image patches of resolution $N \times N$. In this paper we typically use $N = 16$ (“/16”) or $N = 32$ (“/32”). Besides the official DeiT models, we also utilize the data-efficient training scheme to train ViT-L/16 and ViT-B/32 and rename them DeiT-L/16 and DeiT-B/32.

\item \textbf{Big Transfer.} Big Transfer models \cite{DBLP:conf/eccv/KolesnikovBZPYG20} are build on ResNet-V2 models. We select BiT-S-R50X1 based on a ResNet-50 backbone. Besides the official implementation, we also train a version using the identical data augmentation strategy from DeiTs for comparison. We respectively name them BiT and BiT$_{da}$. 
\end{packed_itemize}


\subsection{Evaluation Protocols}

In image classification tasks, a model generally consists of a feature encoder $F$ and a classifier $C$. Suppose the model is trained on a training set $\mathscr{D}_{train}=\{({\rm x}_i,y_i)\}_{i=1}^{N_{train}}$. We respectively introduce a set of independent identically distributed (IID) validation data $\mathscr{D}_{iid}=\{({\rm x}_i,y_i)\}_{i=1}^{N_{iid}}$ and a set of out-of-distributed (OOD) data $\mathscr{D}_{ood}=\{({\rm x}_i,y_i)\}_{i=1}^{N_{ood}}$ in the same semantic space. $N_{train},N_{iid},N_{ood}$ represent the number of data in $\mathscr{D}_{train},\mathscr{D}_{iid},\mathscr{D}_{ood}$ respectively. Then we use the following evaluations.

\begin{packed_itemize}
  \item \textbf{Accuracy on OOD Data.} A direct measurement is to calculate the accuracy on the OOD dataset:
  \begin{equation}
    Acc(F,C;\mathscr{D}_{ood})=\frac{1}{|\mathscr{D}_{ood}|}\sum_{({\rm x},y)\in \mathscr{D}_{ood}} \mathbbm{1}(C(F({\rm x})) = y),
    \label{equa: acc}    
    \end{equation}
     where $\mathbbm{1}$ is the indicator function.
  \item \textbf{IID/OOD Generalization Gap.} In this paper, we also focus on how well a model could behave towards the OOD data compared with the IID data. Hence, we use the IID/OOD generalization gap to measure the performance difference caused by the distribution shift:
    \begin{equation}
    Gap(F,C;\mathscr{D}_{iid},\mathscr{D}_{ood})=Acc(F,C;\mathscr{D}_{iid})-Acc(F,C;\mathscr{D}_{ood}).
    \label{equa: generalization gap OOD}    
    \end{equation}

\end{packed_itemize}

\begin{figure*}[t!]
\centering
\subfloat[T-ADV]{
\includegraphics[width=0.264\linewidth]{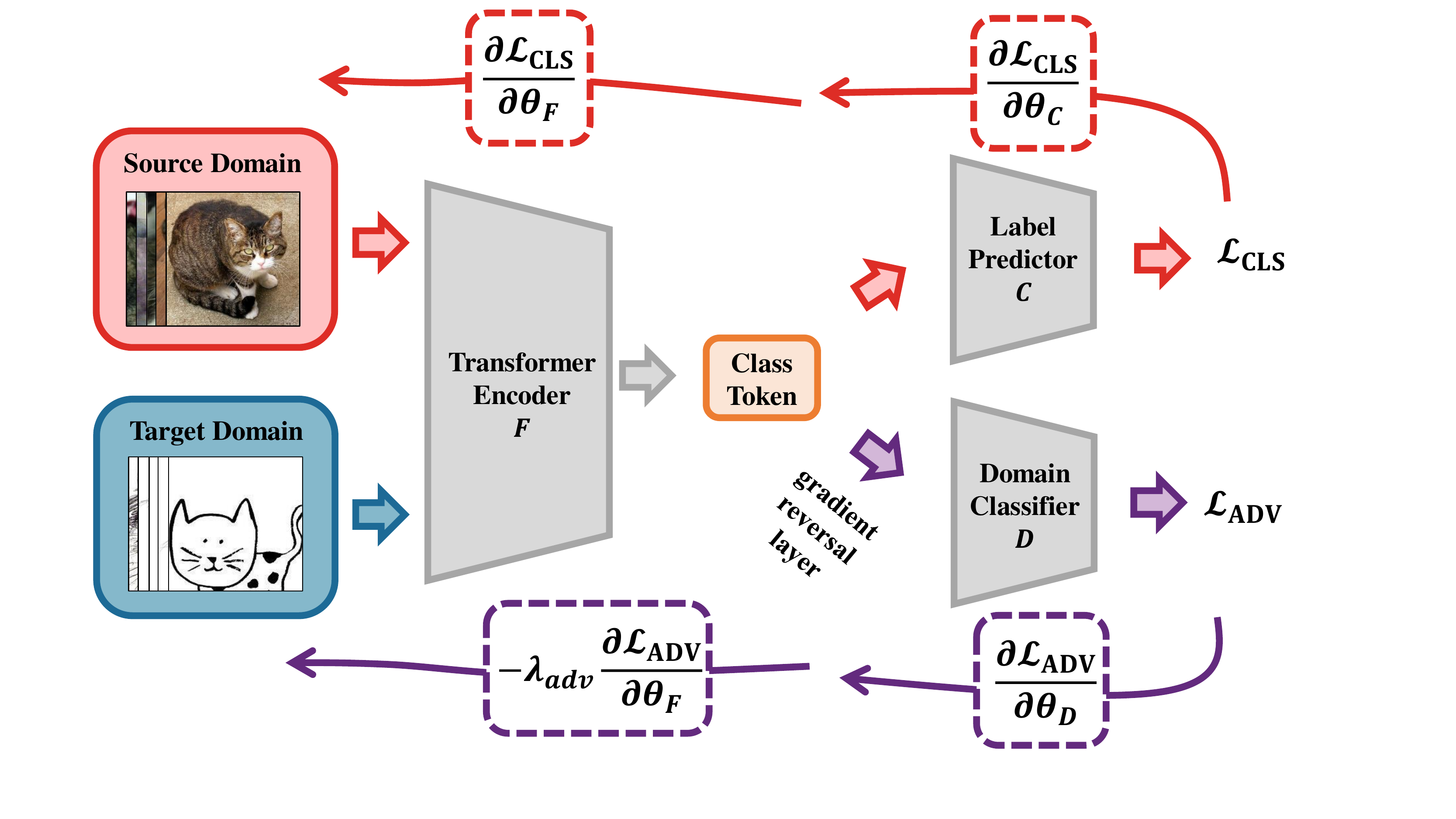}
}
\subfloat[T-MME]{
\includegraphics[width=0.32\linewidth]{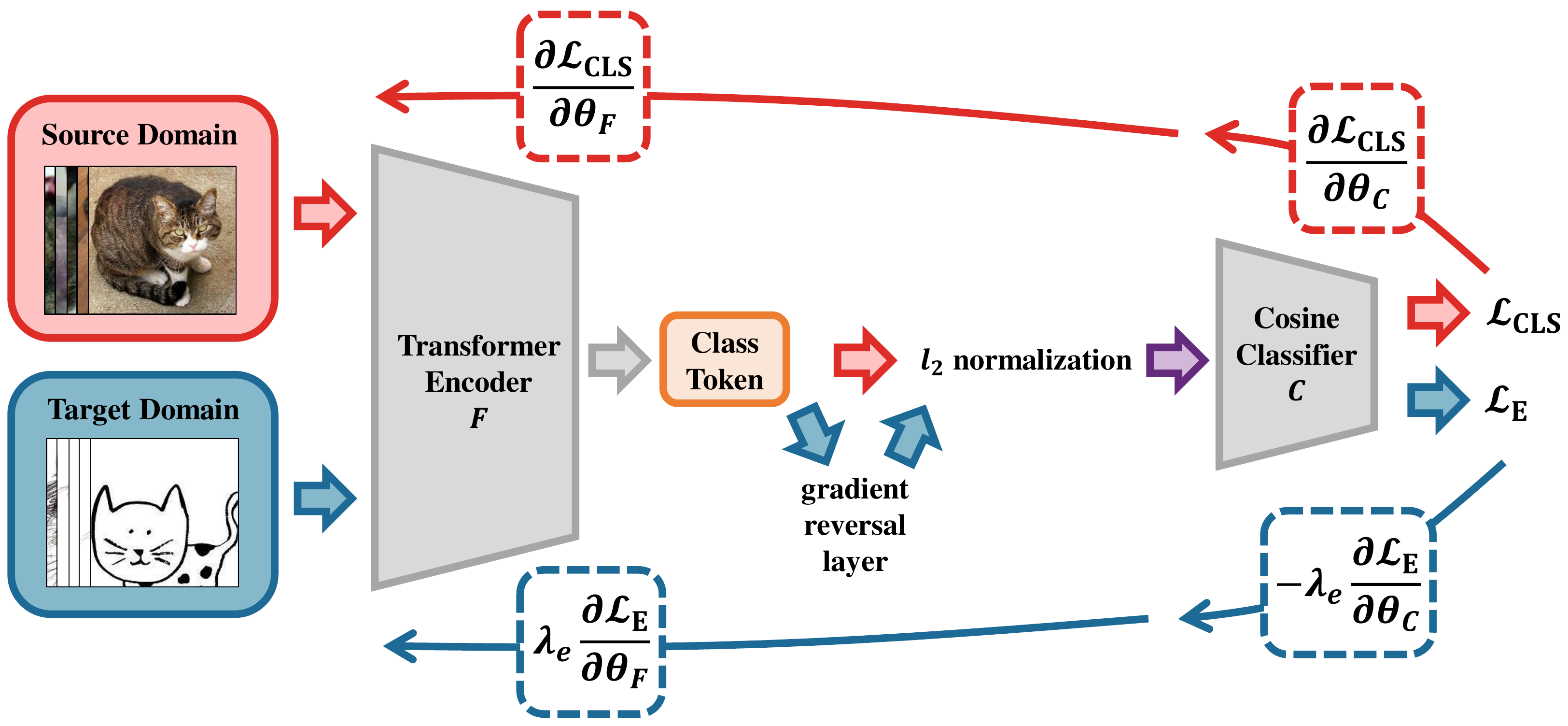}
}\\
\subfloat[T-SSL]{
\includegraphics[width=0.6\linewidth]{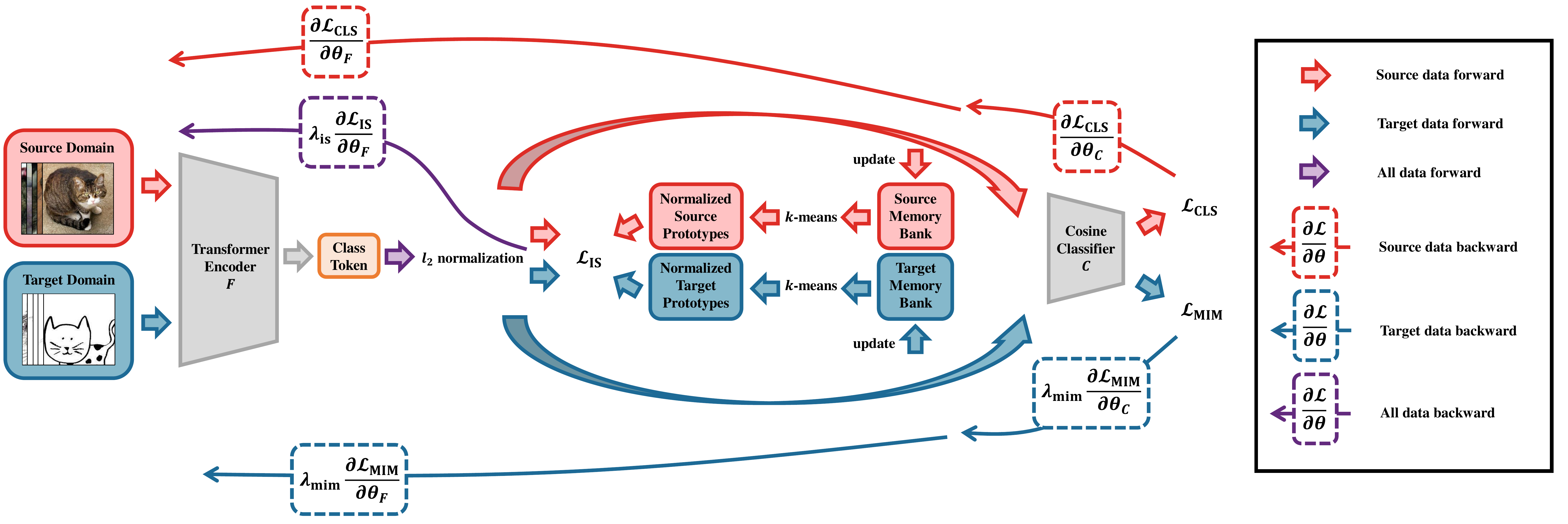}
}
\vspace{-10pt}
\caption{\textbf{A framework overview of the three designed generalization-enhanced ViTs.} All networks use a ViT $F$ as feature encoder and a label prediction head $C$. Under this setting, the inputs to the models have labeled source examples and unlabeled target examples. \textbf{a)} \textbf{T-ADV} promotes the network to learn domain-invariant representations by introducing a domain classifier $D$ for domain adversarial training. \textbf{b)} \textbf{T-MME} leverage the minimax process on the conditional entropy of target data to reduce the distribution gap while learning discriminative features for the task. The network uses a cosine similarity-based classifier architecture $C$ to produce class prototypes. \textbf{c)} \textbf{T-SSL} is an end-to-end prototype-based self-supervised learning framework. The architecture uses two memory banks $V^s$ and $V^t$ to calculate cluster centroids. A cosine classifier $C$ is used for classification in this framework.
}
\vspace{-10pt}
\label{fig:pipeline}
\end{figure*}

\section{Generalization-Enhanced ViTs} \label{sec:TransformerDA}

After investigating the OOD generalization properties of ViTs, it is natural to figure out strategies to further improve them. 
Thus we further design Generalization-Enhanced ViTs (GE-ViTs) from the perspectives of adversarial training \cite{ganin2015unsupervised}, information theory \cite{saito2019semi} and self-supervised learning \cite{yue2021prototypical}, named as T-ADV, T-MME, and T-SSL respectively.
By making a full comparison of these three designs, we figure out the most suitable strategy for GE-ViTs.

\subsection{Adversarial Learning} 

To learn domain-invariant representations, we introduce a domain discriminator \cite{ganin2015unsupervised} to promote the backbone to produce domain-confused features by adversarial training. Specifically, as shown in Fig \ref{fig:pipeline} (a), the network consists of a shared feature encoder $F$, a label predictor $C$, and a domain classifier $D$. The feature encoder aims at minimizing the domain confusion loss $\mathcal{L}_{\mathrm{ADV}}$ for all samples and label prediction loss $\mathcal{L}_{\mathrm{CLS}}$ for labeled source samples while the domain classifier focus on maximizing the domain confusion loss $\mathcal{L}_{\mathrm{ADV}}$. The overall objectives are:
\begin{equation}
    \mathcal{L}_{\mathrm{CLS}} = \sum_{(\mathrm{x},y)\in\mathscr{D}_s} \mathcal{H}(\sigma(C(F(\mathrm{x}))),y),
\end{equation}
\begin{equation}
    \mathcal{L}_{\mathrm{ADV}} = \sum_{(\mathrm{x},y_{d})\in\mathscr{D}_s,\mathscr{D}_t} \mathcal{H}(\sigma(D(F(\mathrm{x}))),y_{d}), 
\end{equation}
\begin{equation}
    (\hat{\theta}_F,\hat{\theta}_C) = \arg \min_{\theta_F,\theta_C} \mathcal{L}_{\mathrm{CLS}} + \lambda_{adv} \mathcal{L}_{\mathrm{ADV}},
\end{equation}
\begin{equation}
    \hat{\theta}_D = \arg \max_{\theta_D} \mathcal{L}_{\mathrm{ADV}}, 
\end{equation}
where $y$ and $y_d$ denote the class label and binary domain label respectively. $\sigma(\cdot)$ stands for the Softmax function and $\mathcal{H}(\cdot,\cdot)$ returns the cross-entropy of two input distributions. $\lambda_{adv}$ is an adaptive coefficient that gradually changed from 0 to 1 by the schedule proposed in \cite{ganin2015unsupervised}. Furthermore, to facilitate training, a gradient reversal layer (GRL) is applied to implement the opposite objective of two parts.

\subsection{Minimax Entropy} \label{sec: mme}

We leverage the minimax process on the conditional entropy of target data \cite{saito2019semi} to reduce the distribution gap while learning discriminative features for the task. As the pipeline is shown in \cref{fig:pipeline} (b), a cosine similarity-based classifier architecture $C$ is exploited to produce class prototypes. The cosine classifier $C$ consists of weight vectors $\mathrm{W} = [\mathrm{w}_1,...,\mathrm{w}_{n_c}]$, where $n_c$ denotes the total number of classes, and a temperature $T$. $C$ takes $\ell_2$ normalized $\frac{F(\mathrm{x})}{\|F(\mathrm{x})\|}$ as an input and output $\frac{1}{T}\frac{\mathrm{W}^{\mathrm{T}} F(\mathrm{x})}{\|F(\mathrm{x})\|}$. The key idea is to minimize the distance between the class prototypes and neighboring unlabeled target samples, thus extracting discriminative target features. To overcome the dominant impact of labeled source data on prototypes, prototypes are moved towards the target by maximizing the entropy $\mathcal{L}_{\mathrm{E}}$ of unlabeled target examples. Meanwhile, the feature extractor aims at minimizing the entropy of the unlabeled examples, to make them better clustered around the prototypes. Therefore, a minimax process is formulated between the weight vectors and the feature extractor. Additionally, the label prediction loss $\mathcal{L}_{\mathrm{CLS}}$ is also utilized on source samples. The overall objectives are:
\begin{equation}
    \mathcal{L}_{\mathrm{CLS}} = \sum_{(\mathrm{x},y)\in\mathscr{D}_s} \mathcal{H}(\sigma(C(F(\mathrm{x}))),y),
\end{equation}
\begin{equation}
    \mathcal{L}_{\mathrm{E}} = \sum_{\mathrm{x}\in\mathscr{D}_t} \mathcal{H}(\sigma(C(F(\mathrm{x})))),
\end{equation}
\begin{equation}
    \hat{\theta}_F = \arg \min_{\theta_F} \mathcal{L}_{\mathrm{CLS}} + \lambda_{e}\mathcal{L}_{\mathrm{E}},
\end{equation}
\begin{equation}
    \hat{\theta}_C = \arg \min_{\theta_C} \mathcal{L}_{\mathrm{CLS}} - \lambda_{e}\mathcal{L}_{\mathrm{E}},
\end{equation}
where $\mathcal{H}(\cdot,\cdot)$ returns the cross-entropy of two input distributions and $\mathcal{H}(\cdot)$ returns the entropy. $\lambda_{e}$ is a coefficient to balance two loss terms.

\begin{figure*}[t]
\centering
\subfloat[]{
\includegraphics[width=0.23\linewidth]{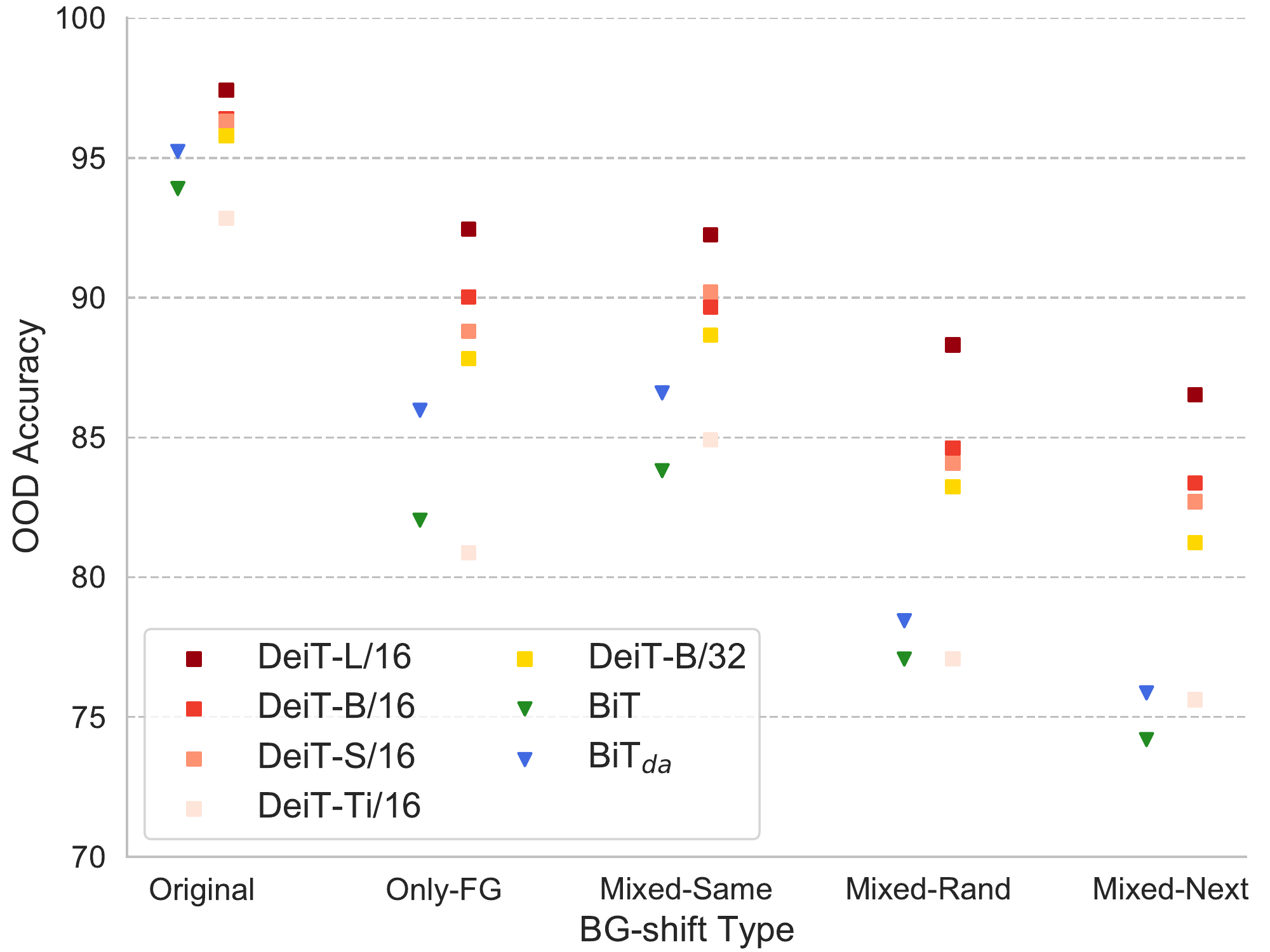}
}
\subfloat[]{
\includegraphics[width=0.23\linewidth]{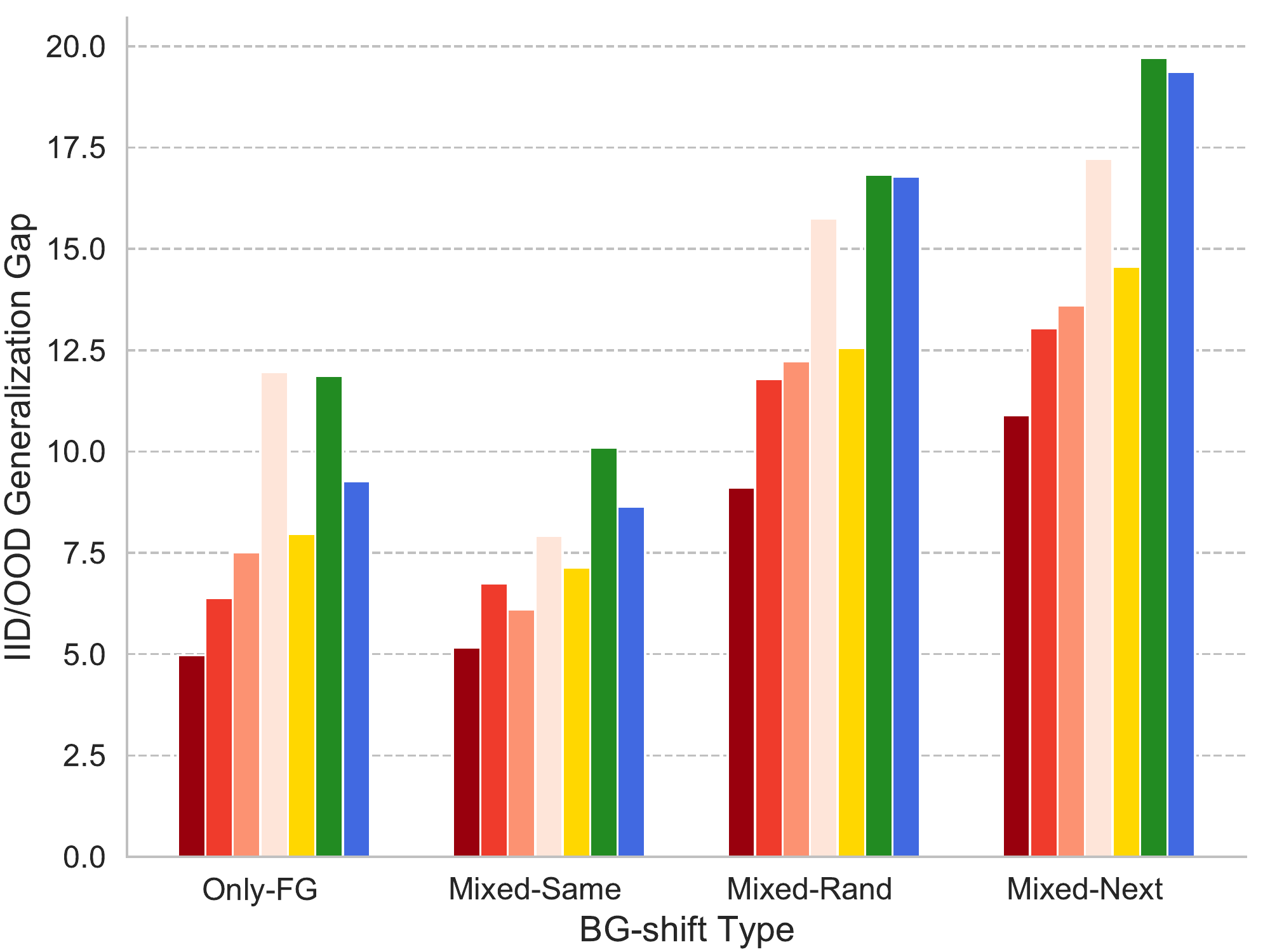}
}
\subfloat[]{
\includegraphics[width=0.23\linewidth]{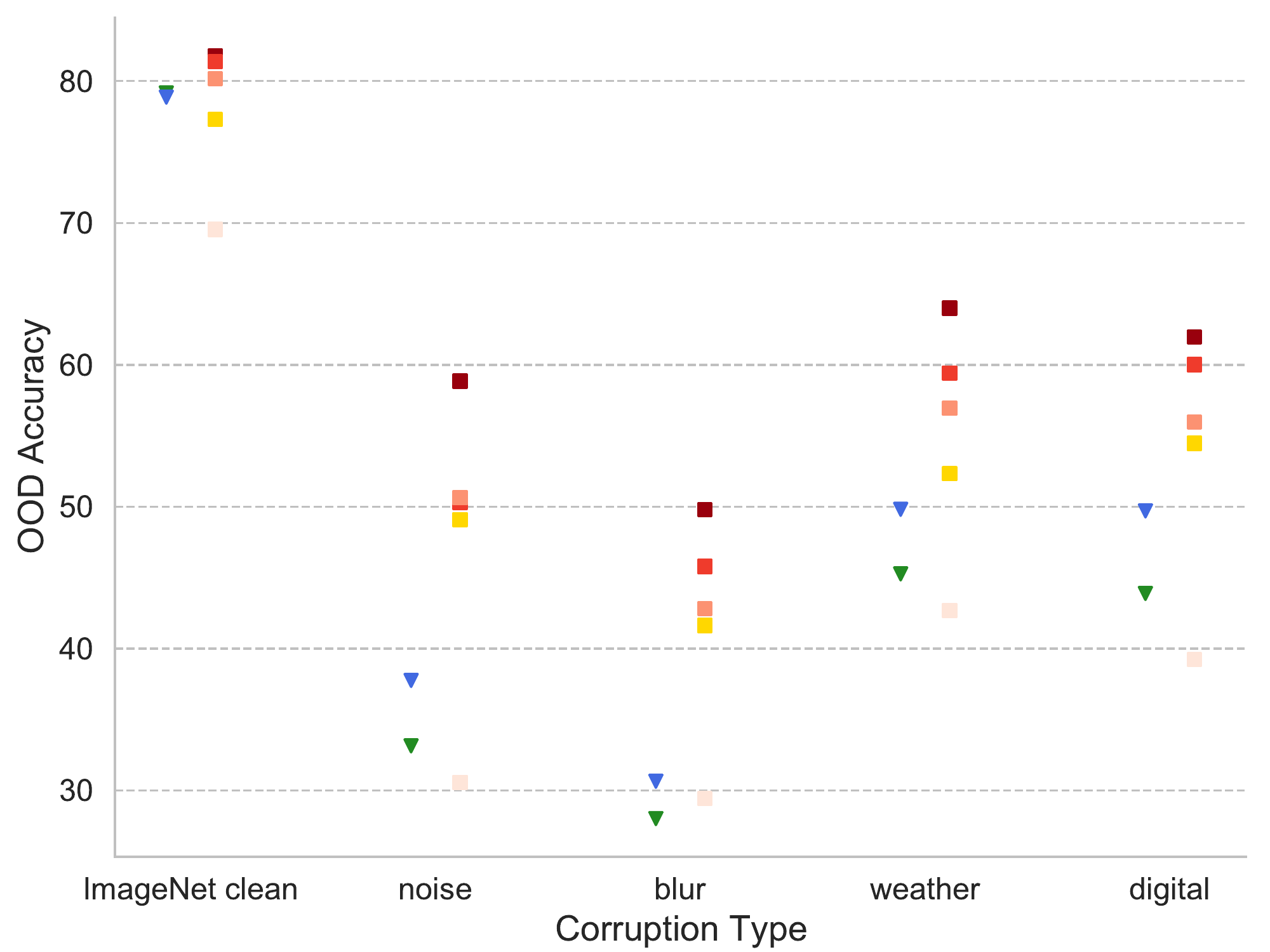}
}
\subfloat[]{
\includegraphics[width=0.23\linewidth]{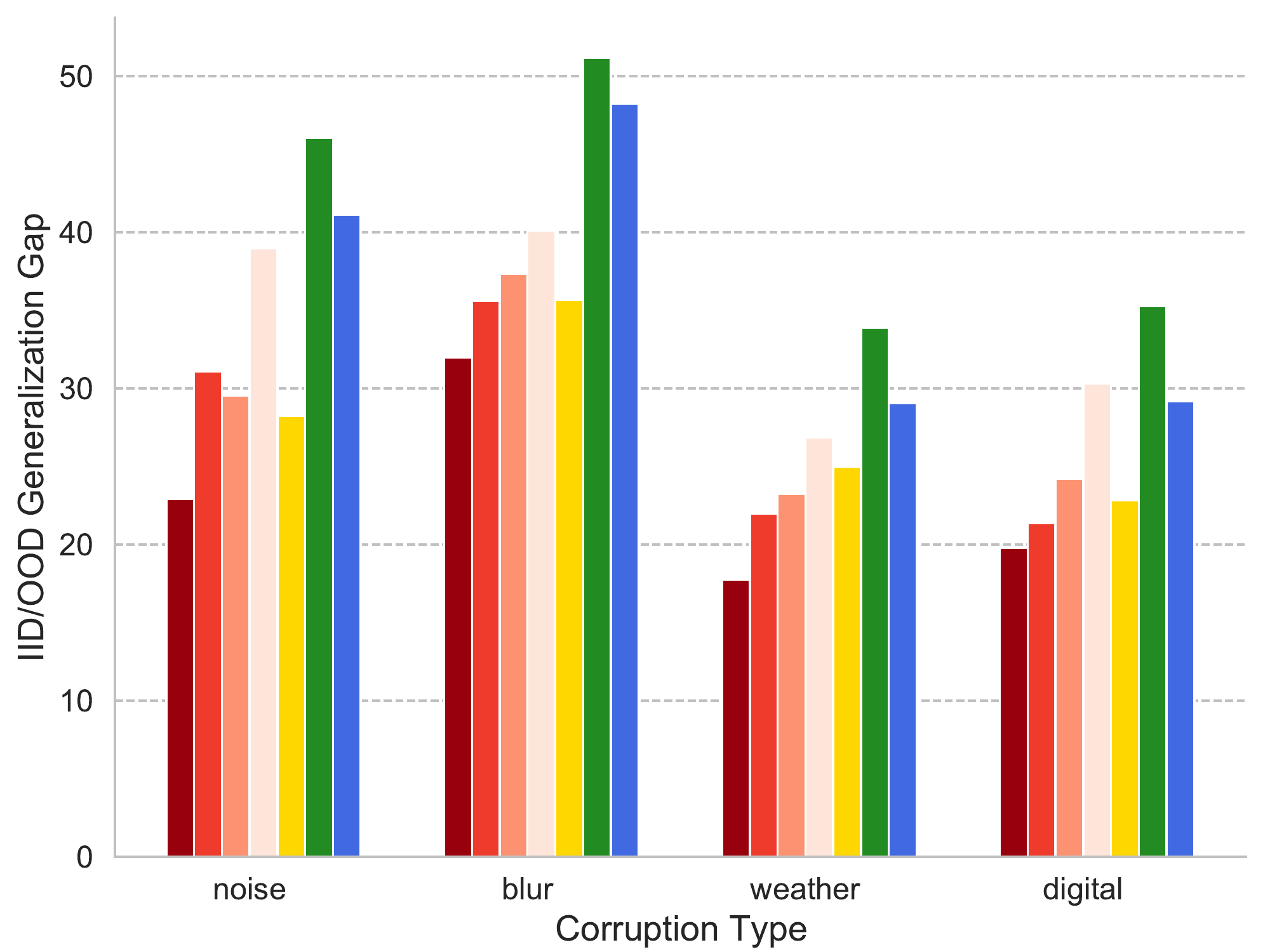}
}
\vspace{-8pt}
\caption{\textbf{Results on ImageNet-9 and ImageNet-C.} (a)-(b) and (c)-(d) respectively illustrate the OOD Accuracy and IID/OOD Generalization Gap for different models on ImageNet-9 and ImageNet-C datasets. From (a) and (b), we conclude that \textbf{1)} ViTs perform with a weaker background-bias than CNNs, \textbf{2)} a larger ViT extracts a more background-irrelevant representation. From (c) and (d), we draw the conclusions that \textbf{1)} ViTs deal with corruption shifts better than CNNs and generalize better along with model size scaling up, \textbf{2)} ViTs do benefit from diverse augmentation in enhancing generalization towards vicinal impurities, but their architectural advantage cannot be overlooked as well, \textbf{3)} patch size for training has little influence on ViTs' generalization ability.}
\vspace{-10pt}
\label{fig: BG_CR}
\end{figure*}

\subsection{Self-Supervised Learning} 


We integrate an end-to-end prototypical self-supervised learning framework \cite{yue2021prototypical} into ViT. As shown in \cref{fig:pipeline} (c), the framework also uses a cosine classifier $C$ as introduced in \cref{sec: mme}. It first encodes semantic structure of data into the embedding space. ProtoNCE \cite{li2020prototypical} is respectively applied in source and target domains. Specifically, two memory banks $V^s$ and $V^t$ are maintained to store feature vectors of every sample from source and target. These vectors are updated with momentum after each batch. $k$-means clustering is performed on memory banks to generate normalized prototypes $\{\mu_j^s\}_{j=1}^k$ and $\{\mu_j^t\}_{j=1}^k$. Then the similarity distribution vector between $\ell_2$ normalized source feature vectors $f_i^s = \frac{F(\mathrm{x}_i^s)}{\|F(\mathrm{x}_i^s)\|}$ from current batch and normalized source prototypes $\{\mu_j^s\}_{j=1}^k$ as $P_i^s = [P_{i,1}^s,...,P_{i,k}^s]$ with $P_{i,j}^s = \frac{\mathrm{exp}(\mu_j^s \cdot f_i^s / \phi)}{\sum_{r=1}^{k} \mathrm{exp}(\mu_r^s \cdot f_i^s / \phi)}$,
%
where $\phi$ is a temperature value. Then the in-domain prototypical self-supervision loss is formed as: $\mathcal{L}_{\mathrm{IS}} = \sum_{i=1}^{|\mathscr{D}_s|} \mathcal{H}(P_i^s,c_s(i))+\sum_{i=1}^{|\mathscr{D}_t|}\mathcal{H}(P_i^t,c_t(i))$, 
where $c_s(\cdot)$ and $c_t(\cdot)$ return the cluster index of the sample, and $|\cdot|$ returns the cardinal of the set. $\mathcal{H}(\cdot,\cdot)$ returns the cross-entropy of two input distributions.




In addition, since a network is desired to have high-confident and diversified predictions, an objective is set for maximizing the mutual information between the input image and the network prediction. This objective is split into two terms: entropy maximization of expected network prediction and entropy minimization on the network output. Therefore, the objective is formulated as: $\mathcal{L}_{\mathrm{MIM}} = \mathbb{E}_{\mathrm{x}}[\mathcal{H}(p(y|\mathrm{x};\theta)] - \mathcal{H}(\mathbb{E}_{\mathrm{x}\in\mathscr{D}_s \cup \mathscr{D}_t}[p(y|\mathrm{x};\theta])$.
%
The last term of training objective is the supervision loss on source domain measured by cross-entropy: $\mathcal{L}_{\mathrm{CLS}} = \sum_{(\mathrm{x},y)\in\mathscr{D}_s} \mathcal{H}(\sigma(C(F(\mathrm{x}))),y)$.

Finally, the overall learning objective is formulated as:
\begin{equation}
(\hat{\theta}_F, \hat{\theta}_C) = \arg \min_{\theta_F,\theta_C} \mathcal{L}_{\mathrm{CLS}} + \lambda_{\mathrm{is}} \mathcal{L}_{\mathrm{IS}} + 
\lambda_{\mathrm{mim}} \mathcal{L}_{\mathrm{MIM}},
\label{equa: PCS objective}
\end{equation}
where $\lambda_{\mathrm{is}}$ and $\lambda_{\mathrm{mim}}$ denotes the coefficients of corresponding loss terms.

\section{Systematic Study on ViTs Generalization}


\noindent\textbf{In-Distribution Generalization.}
We first examine the in-distribution generalization of different models on the ImageNet benchmark. As results are shown in \cref{fig: BG_CR} (c) column 1, we have the following observations. \textbf{1)} With the data-efficient training scheme, DeiT models tend to perform better as scales increase from \emph{tiny} to \emph{large}, but the gain of scale growth gradually dwindles. \textbf{2)} Having almost the same parameters and both trained without external data, DeiT-S/16 could beat BiT and BiT$_{da}$.


\subsection{Background Shifts Generalization Analysis}

We utilize ImageNet-9, a variety of foreground-background recombination plans, to investigate model bias towards background signal. 
These datasets empower us to investigate to what extent model decisions rely on the background signal. The OOD accuracy and IID/OOD gap results of four varieties of background shifts are illustrated in \cref{fig: BG_CR} (a) and (b) respectively.


\noindent\textbf{- ViTs perform with a weaker background-bias than CNNs.} 
By calculating accuracy gaps between \emph{Mixed-Same} with class-relevant backgrounds and \emph{Mixed-Rand} with neutral background signals, we can measure classifiers' reliance on the correct background. 
From \cref{fig: BG_CR} (a), the lower gaps achieved by ViTs indicate that ViTs depend less on corresponding background signals when the correct foreground is present. Likewise, it can be concluded that ViTs are less misled by the conflict background based on the accuracy gaps between \emph{Mixed-Same} and \emph{Mixed-Next}. In addition, comparing two BiT models, BiT$_{da}$ outperforms normal BiT in OOD accuracy and achieves lower IID/OOD Gaps, indicating that diverse augmentation during training exerts a salutary effect on model generalization on background-shifted data. Nonetheless, it is worth noticing that BiT$_{da}$ obtain a larger \emph{Same-Rand} gap and \emph{Same-Next} gap, which demonstrates that the augmentation training scheme cannot alleviate the model's dependence on correct background information. Therefore, ViTs perform with a weaker background bias than CNNs, and such property is brought by their architectures.


\begin{figure*}[t]
\centering
\subfloat[Stylized-ImageNet]{
\includegraphics[width=0.23\linewidth]{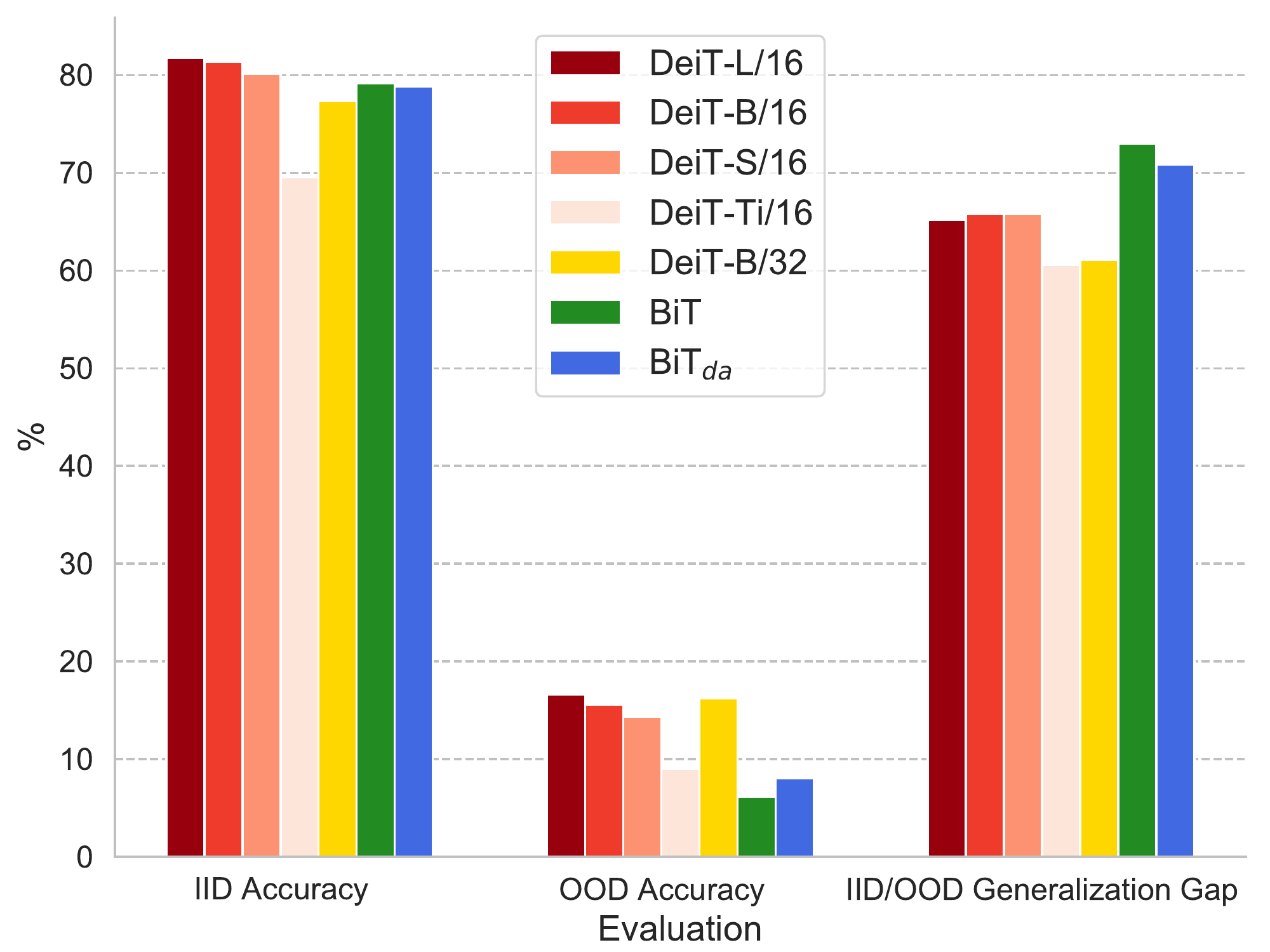}
}
\hspace{0.02\linewidth}
\subfloat[Cue Conflict Stimuli]{
\includegraphics[width=0.23\linewidth]{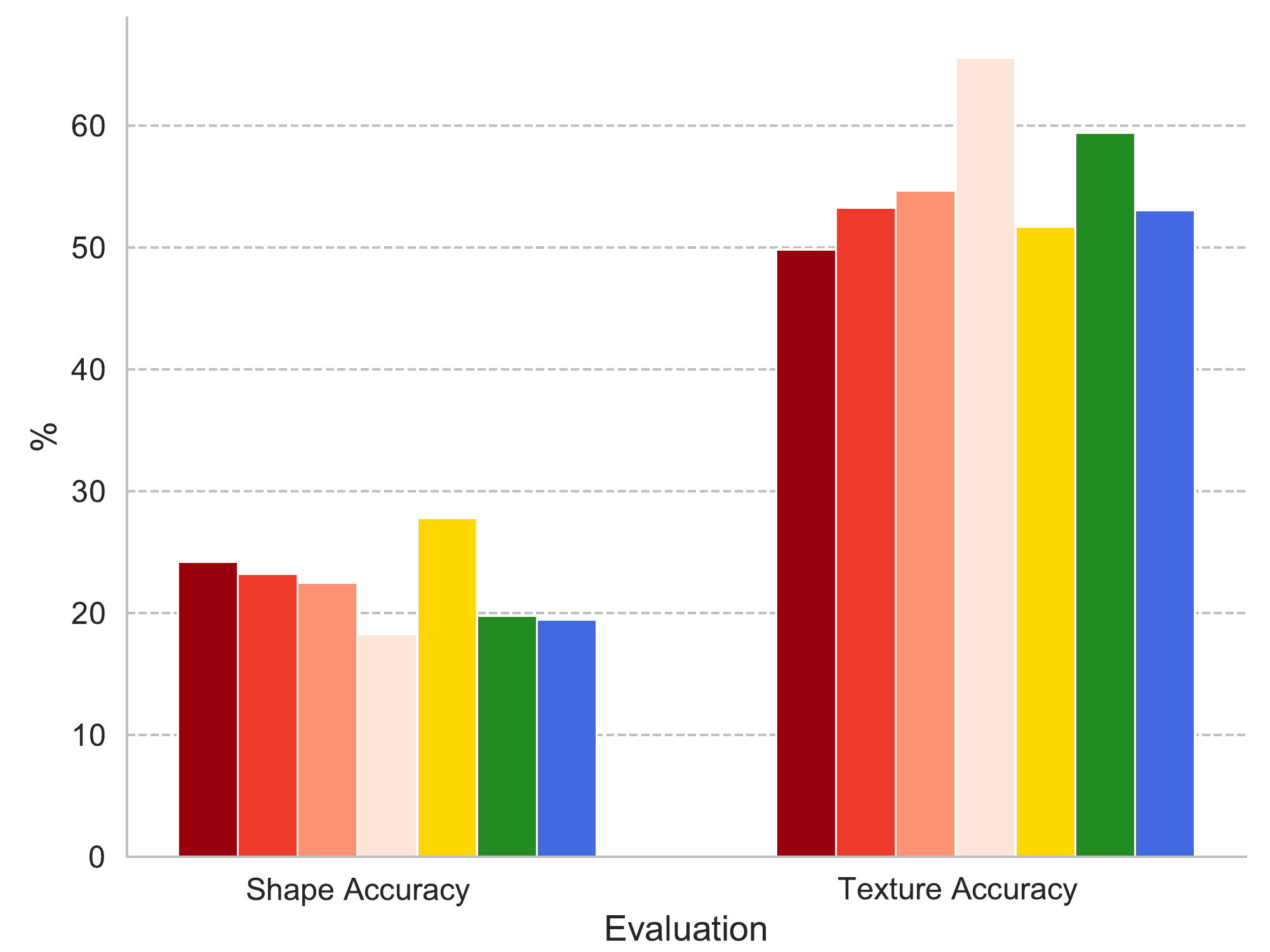}
}
\hspace{0.02\linewidth}
\subfloat[ImageNet-R]{
\includegraphics[width=0.23\linewidth]{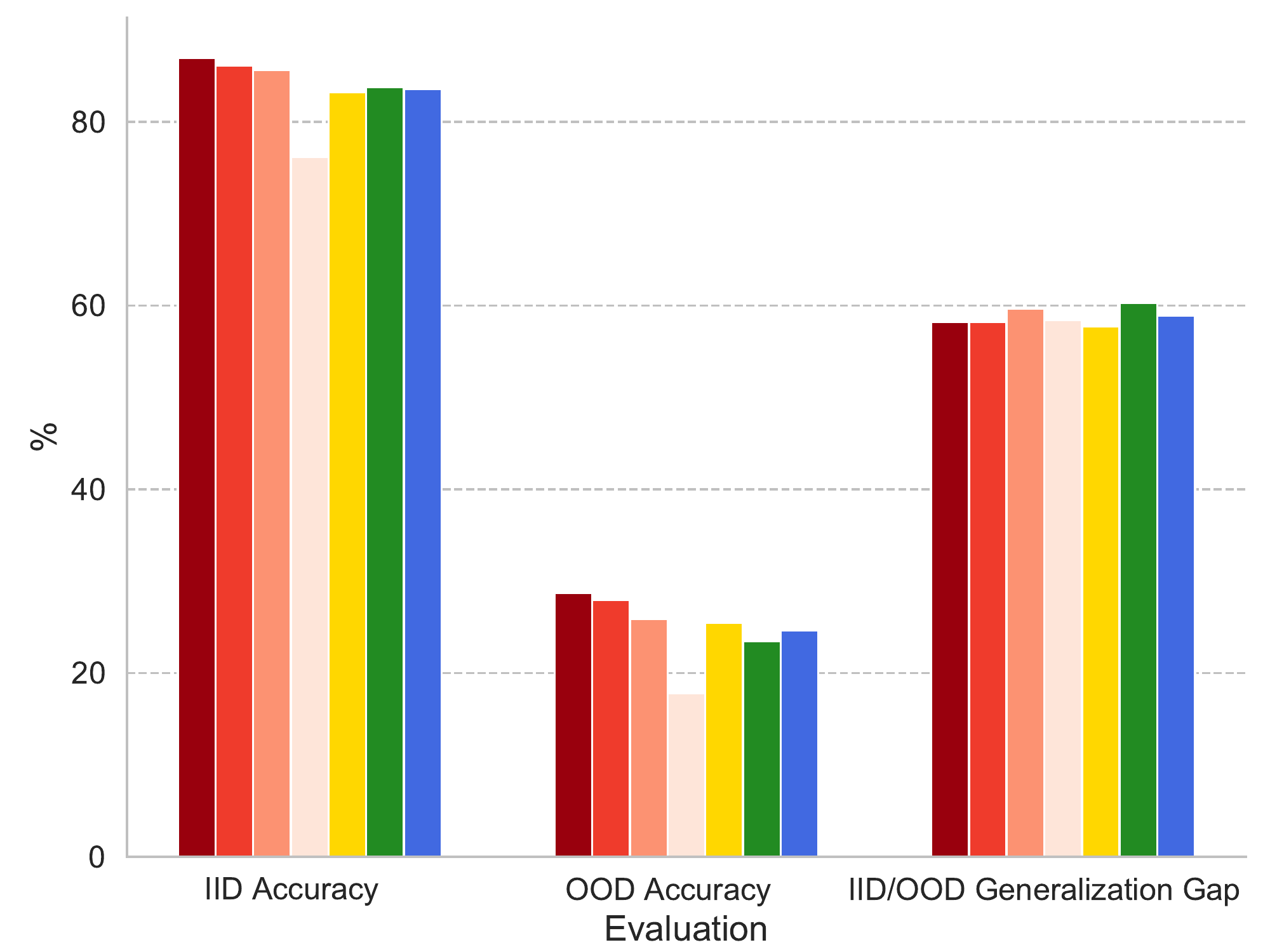}
}
\vspace{-8pt}
\caption{\textbf{Results on Stylized-ImageNet, Cue Conflict Stimuli and ImageNet-R.} (a), (b) and (c) respectively illustrate the OOD Accuracy and IID/OOD Generalization Gap for different models on Stylized-ImageNet, Cue Conflict Stimuli, and ImageNet-R data sets. From (a) and (b) we could draw the following conclusions that \textbf{1)} ViTs' stronger bias towards shape enables them to generalize better under texture shifts and their shape biases have a positive correlation with their sizes, \textbf{3)} ViTs with larger patch size exhibit a stronger bias towards the shape. From (c) we observe that most ViTs beat BiTs in OOD accuracy while having little difference in the IID/OOD generalization gap.}
\label{fig: TX_STY}
\vspace{-10pt}
\end{figure*}

\noindent\textbf{- A larger ViT extracts a more background-irrelevant representation.} Via comparing ViTs of different sizes, we can observe that a larger ViT architecture contributes to a better OOD performance as well as a smaller IID/OOD gap. Even DeiT-L/16 could further narrow the gap by about 2\% from DeiT-B/16, while they achieve almost the same in distribution accuracy results. Meanwhile, a larger ViT also achieves a lower \emph{Same-Rand} gap and \emph{Same-Next} gap, showing that there exists a positive correlation between the ViT scale and their ability to exclude distraction provided by irrelevant or conflict backgrounds. Hence, it is clear that a larger ViT tends to focus more attention on the foreground and learn a more background-irrelevant representation.
                        


\begin{figure*}[t]
\centering
\subfloat[DeiT-B/16]{
\includegraphics[width=0.23\linewidth]{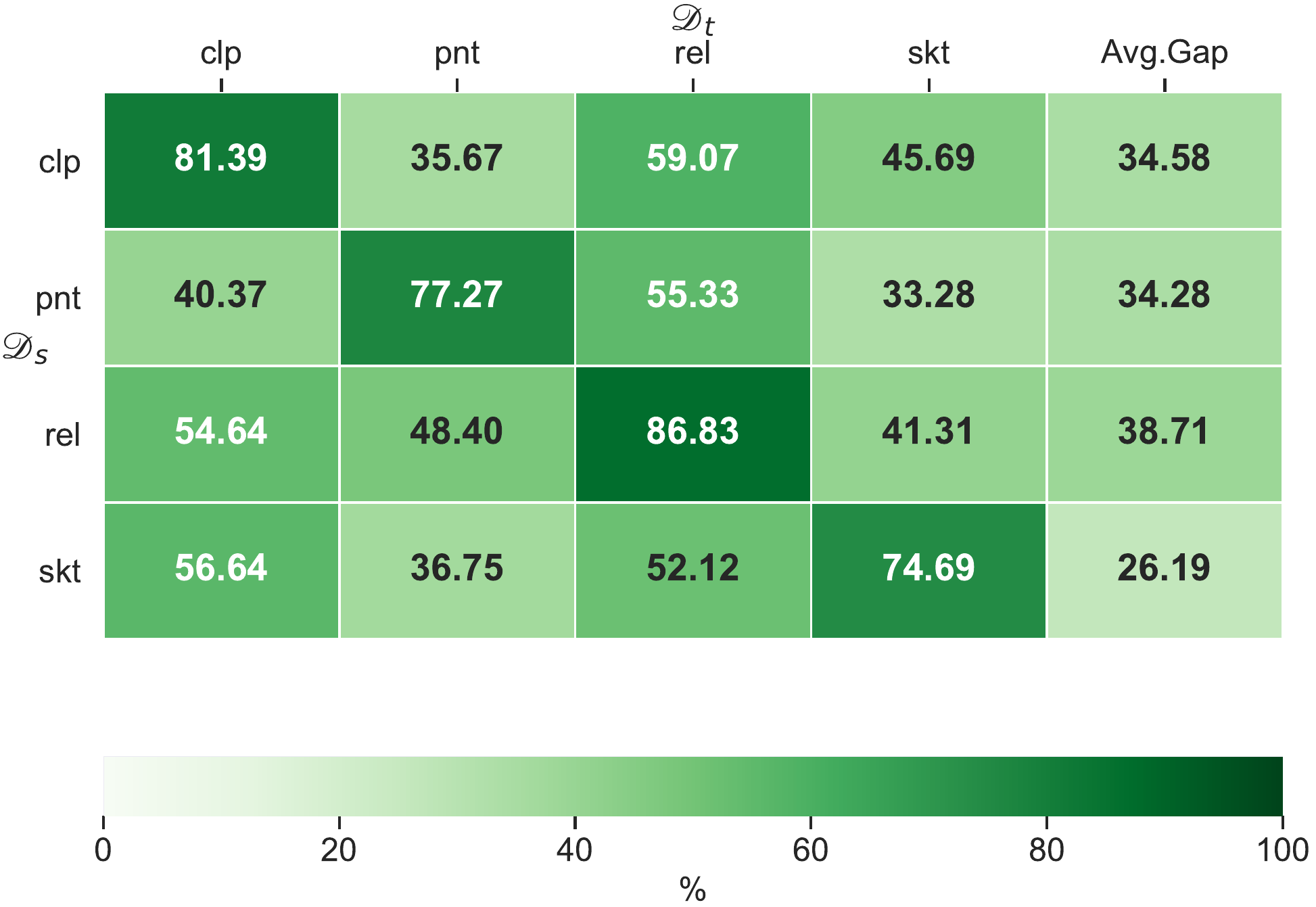}
}
\subfloat[DeiT-S/16]{
\includegraphics[width=0.23\linewidth]{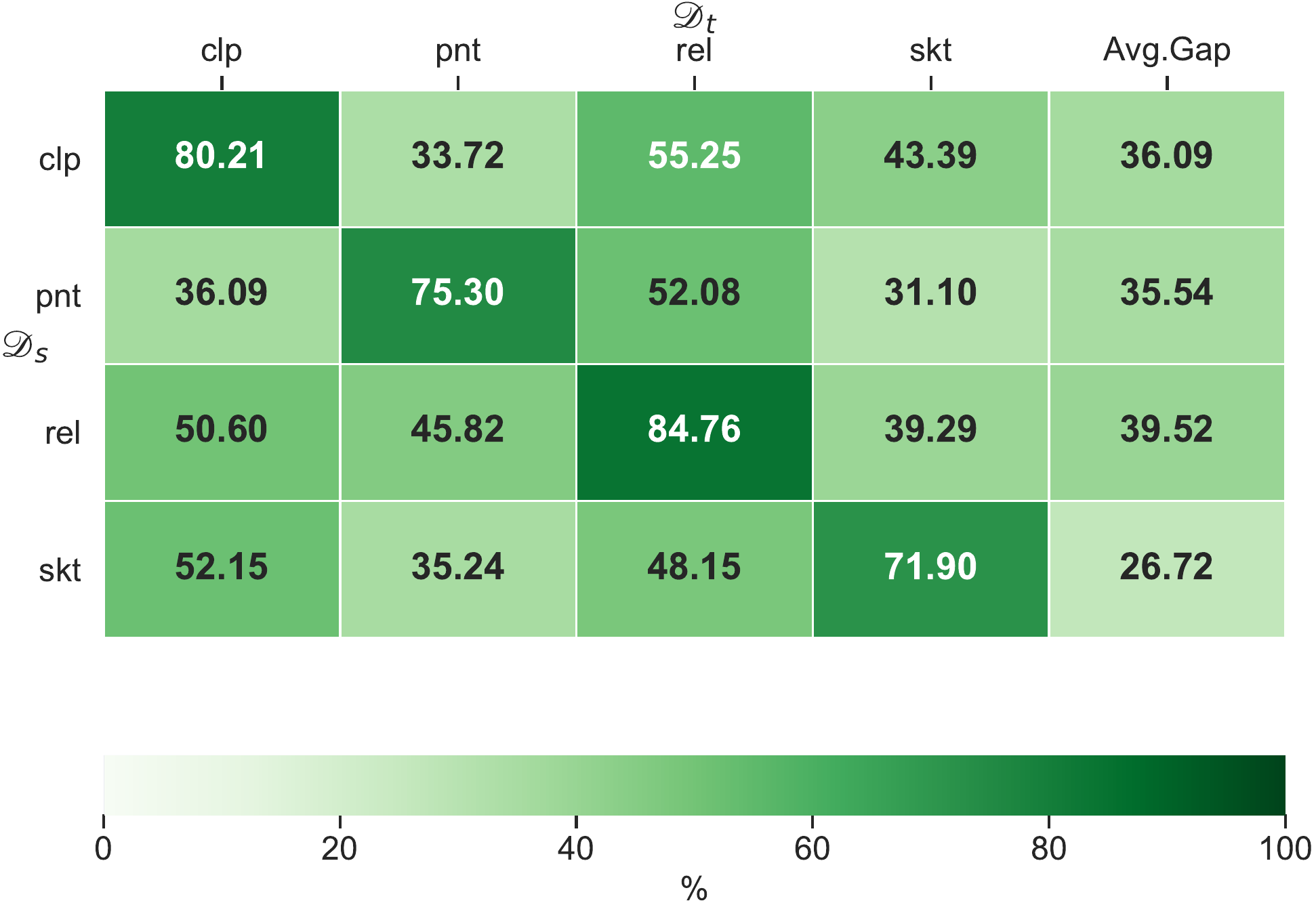}
}
\subfloat[BiT]{
\includegraphics[width=0.23\linewidth]{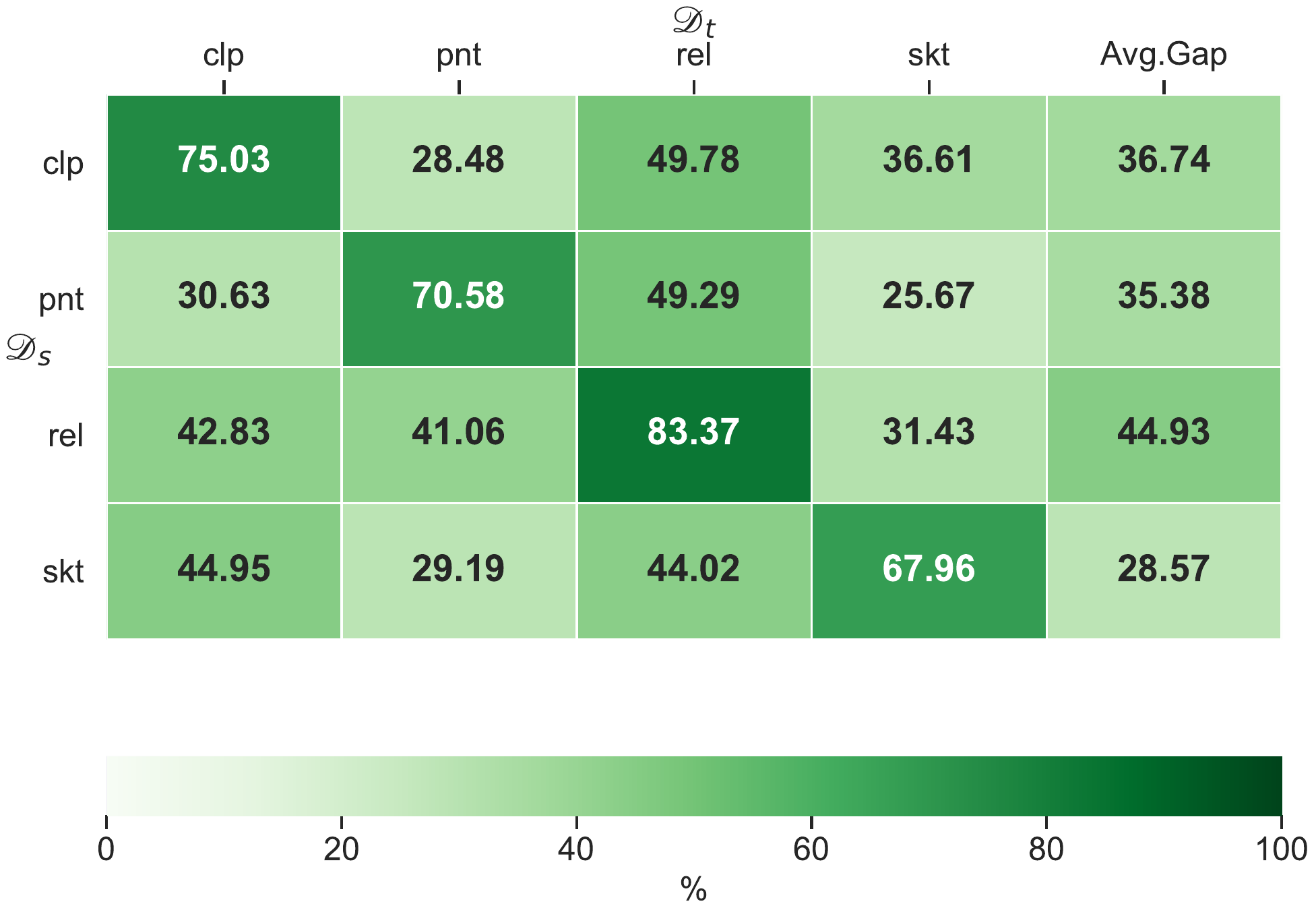}
}
\subfloat[BiT$_{da}$]{
\includegraphics[width=0.23\linewidth]{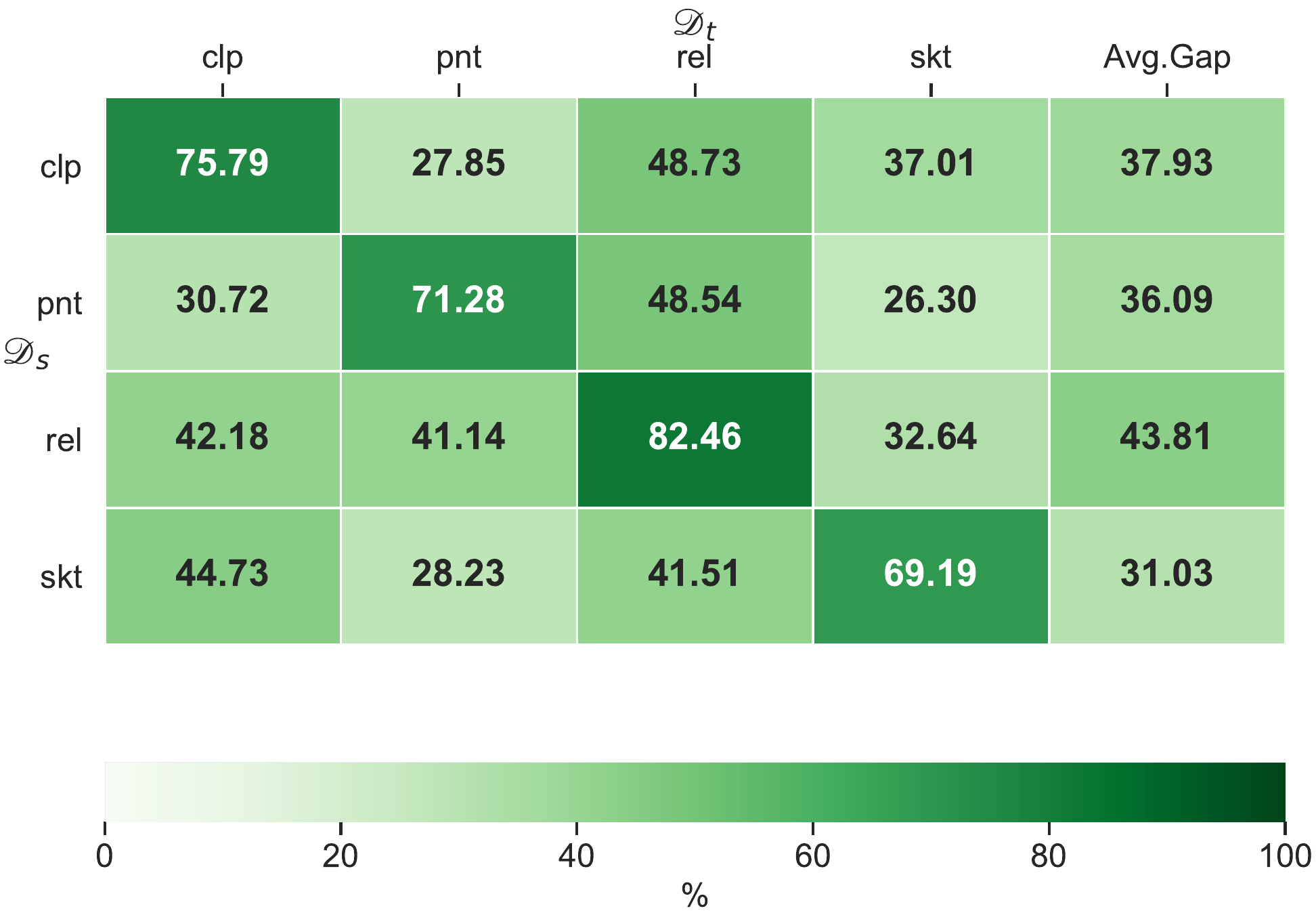}
}
\vspace{-8pt}
\caption{\textbf{Results on DomainNet.}  From the results, we can conclude that \textbf{1)} DeiT-S/16 performs better on the small-scale datasets in IID conditions. Thus, the model easily outperforms BiTs in OOD accuracy, \textbf{2)} when inspecting the IID/OOD generalization gap, the results differ a lot. When models are trained on clipart and painting, there is no obvious difference of gap between DeiT-S/16 and BiTs.}
\label{fig: domainnet}
\vspace{-10pt}
\end{figure*}

\begin{figure}[t]
\centering
\subfloat[]{
\includegraphics[width=0.8\linewidth]{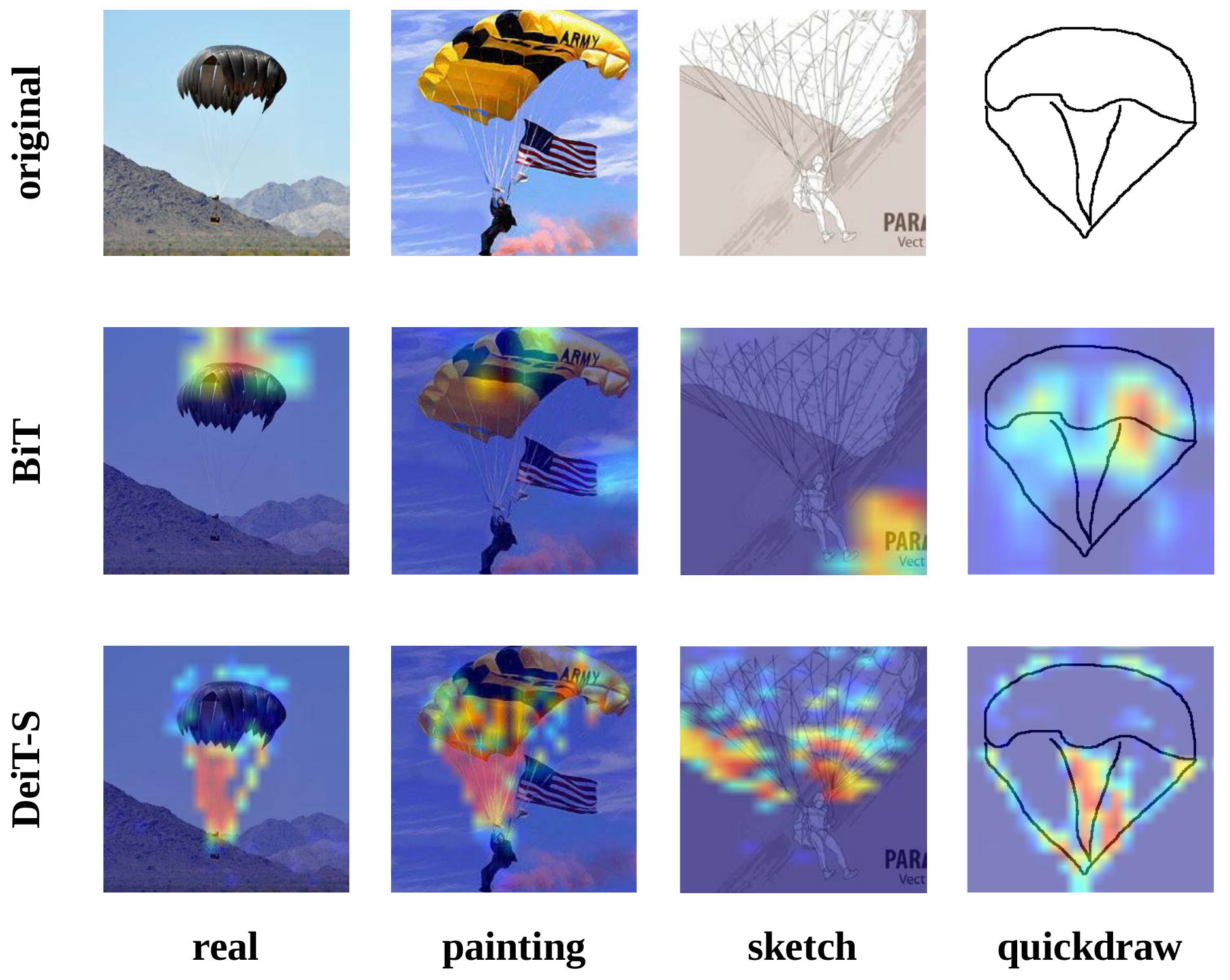}
}\\
\subfloat[]{
\includegraphics[width=0.5\linewidth]{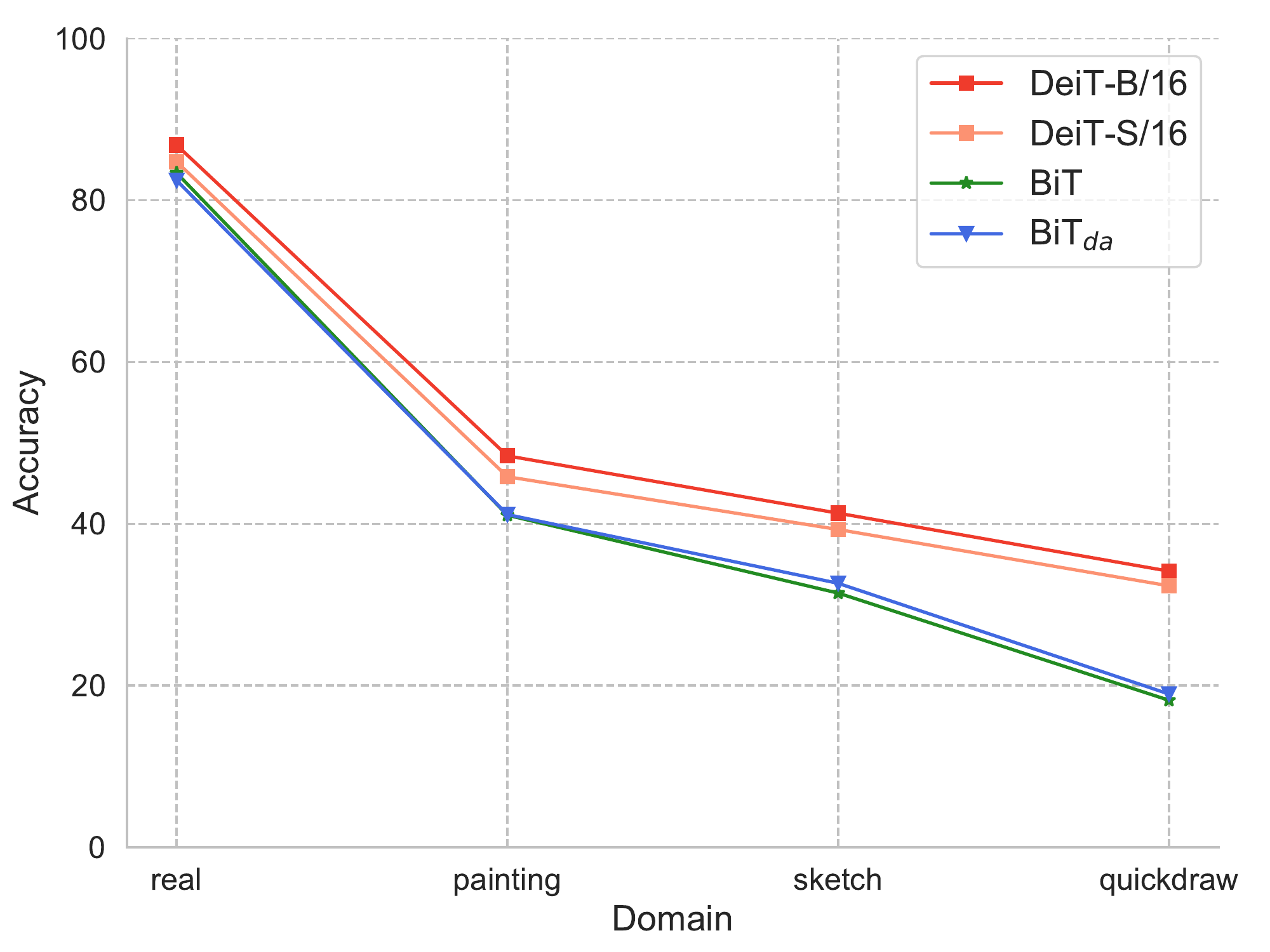}
}
\vspace{-10pt}
\caption{\textbf{Structure bias investigation.} (a) illustrates examples of class \emph{parachute} of four domains and the Grad-CAM \cite{selvaraju2017grad} attention maps of both BiT and DeiT-S. We shall observe that, as the color, texture, and shape cues become less and less informative from \emph{real} to \emph{quickdraw} and even there is only abstract structure preserved in \emph{quickdraw}, DeiT-S constantly concentrates on the key structural information of parachutes while BiT fails to capture such essential feature. (b) shows the accuracies of models trained with real on different domains. From the results, we can see that the gap between ViTs and CNNs is getting larger when the tested domain contains fewer visual cues (\ie from real to quickdraw). Therefore, we can conclude that ViTs are less affected by the shift of color, texture, and shape features, indicating that ViTs focus more on structures.}
\label{fig: diff_domain_demo}
\vspace{-10pt}
\end{figure}

\subsection{Corruption Shifts Generalization Analysis}

The corruption results of 4 categories averaged over all subclasses and all severities, are shown in \cref{fig: BG_CR} (c) and (d). \\
\textbf{- ViTs deal with corruption shifts better than CNNs and generalize better along with model size scaling up.} 
There exist similar phenomena with the background shifts cases that most ViTs lead the BiT models to a large extent under both evaluations in all situations, and that a larger ViT architecture achieves a better OOD performance and narrows the IID/OOD generalization gap.\\
\textbf{- ViTs benefit from diverse augmentation in enhancing generalization towards vicinal impurities, but their architectural advantage cannot be overlooked.} Compared with BiT, BiT$_{da}$ constantly achieves about 4\% better in OOD performances and IID/OOD gaps, emphasizing the contribution of diverse augmentation to model insensitivity towards pixel-level shifts. However, most ViT models are still ahead of BiT$_{da}$ under both evaluations, manifesting ViTs' performance can be partially attributed to the architecture design.\\
\textbf{- Patch size for training has little influence on ViTs' generalization ability.}  Though DeiT-B/16 achieves higher OOD accuracy than DeiT-B/32, its counterpart trained with a larger patch size $32 \times 32$, there is little difference between their IID/OOD gaps. Therefore, patch size for training exert a peripheral effect on generalization ability from in-distribution data to out-of-distribution data, but only act on the model in-distribution generalization.
 

\subsection{Texture Shifts Generalization Analysis}
The results on Stylized-ImageNet and Cue Conflict Stimuli are shown in \cref{fig: TX_STY} (a) and (b). 

\noindent\textbf{- ViTs' stronger bias towards shape enables them to generalize better under texture shifts and their shape biases have a positive correlation with their sizes.} It could be observed from results on Stylized-ImageNet that ViTs lead BiT models under both evaluations and a larger ViT architecture achieves a better OOD performance, which indicates that ViTs deal with the texture shifts better and a larger ViT contributes to better leveraging global semantic features (such as shape and object parts) and less affected by local changes. These phenomena reappear in results on Cue Conflict Stimuli that most ViTs achieve higher shape accuracy and lower texture accuracy than BiTs, which demonstrates that ViTs' insensitivities towards texture shifts are owed to their stronger bias on shape than CNNs. Meanwhile, there exists an uptrend of shape accuracy and a downtrend of texture accuracy as the ViT size increases. Hence, ViTs' shape biases have a positive correlation with their sizes.

\noindent\textbf{- ViTs with larger patch size exhibit a stronger bias towards the shape.} On Stylized-ImageNet, DeiT-B/32 behaves better than DeiT-B/16 in OOD accuracy and IID/OOD generalization gap, which is opposite to their performances on ImageNet. Meanwhile, on Cue Conflict Stimuli, DeiT-B/32 is less affected by the misleading texture than DeiT-B/16, resulting in a higher shape accuracy. Therefore, ViTs with larger patch size rely less on local texture features and focus more on global high-level features, \ie they show stronger bias towards shape lower bias towards texture.

\begin{figure}[t]
\centering
\subfloat[rel vs. pnt, L8]{
\includegraphics[width=0.33\linewidth]{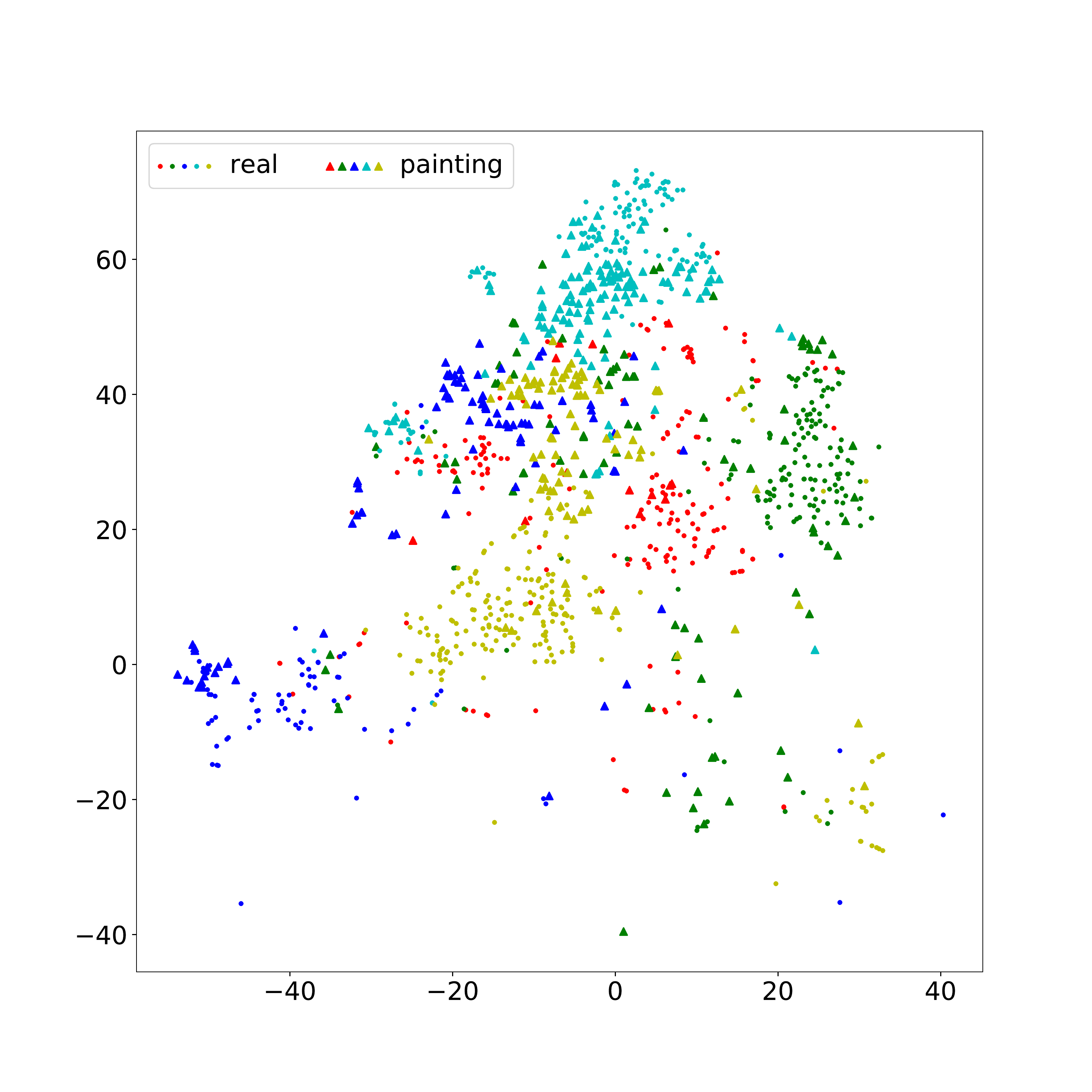}
}
\subfloat[rel vs. skt, L8]{
\includegraphics[width=0.33\linewidth]{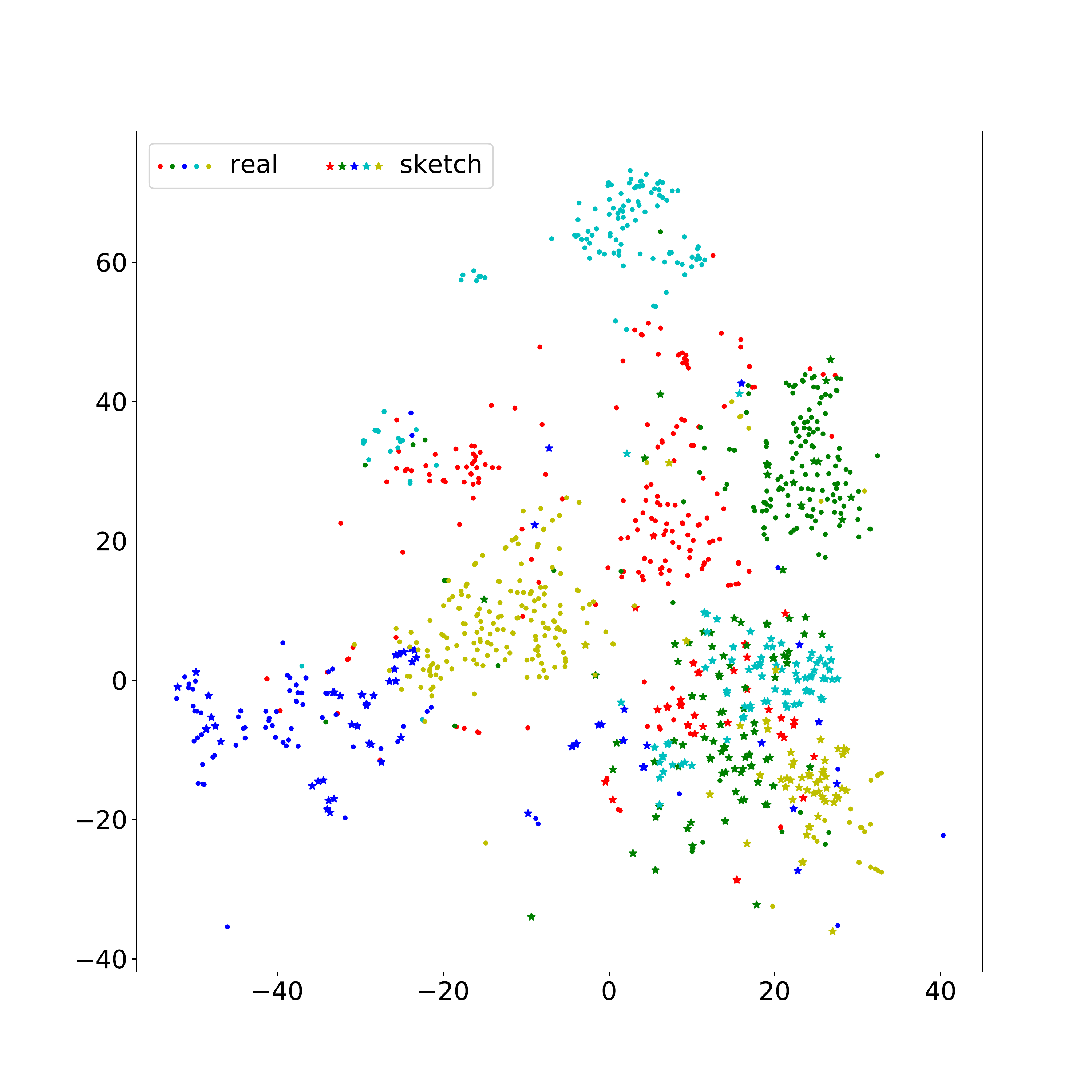}
}
\subfloat[rel vs. qcd, L8]{
\includegraphics[width=0.33\linewidth]{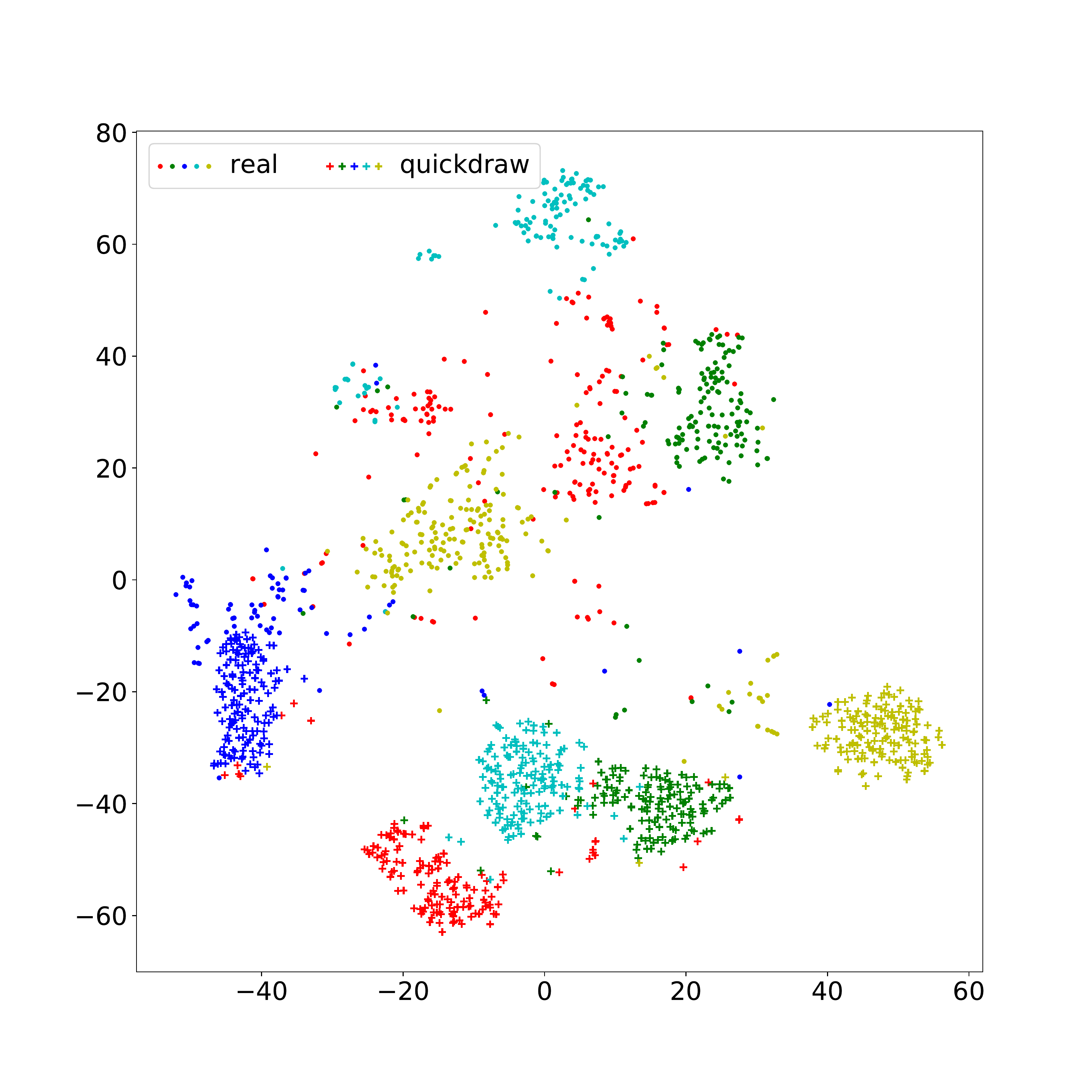}
}\\
\subfloat[rel vs. pnt, L12]{
\includegraphics[width=0.33\linewidth]{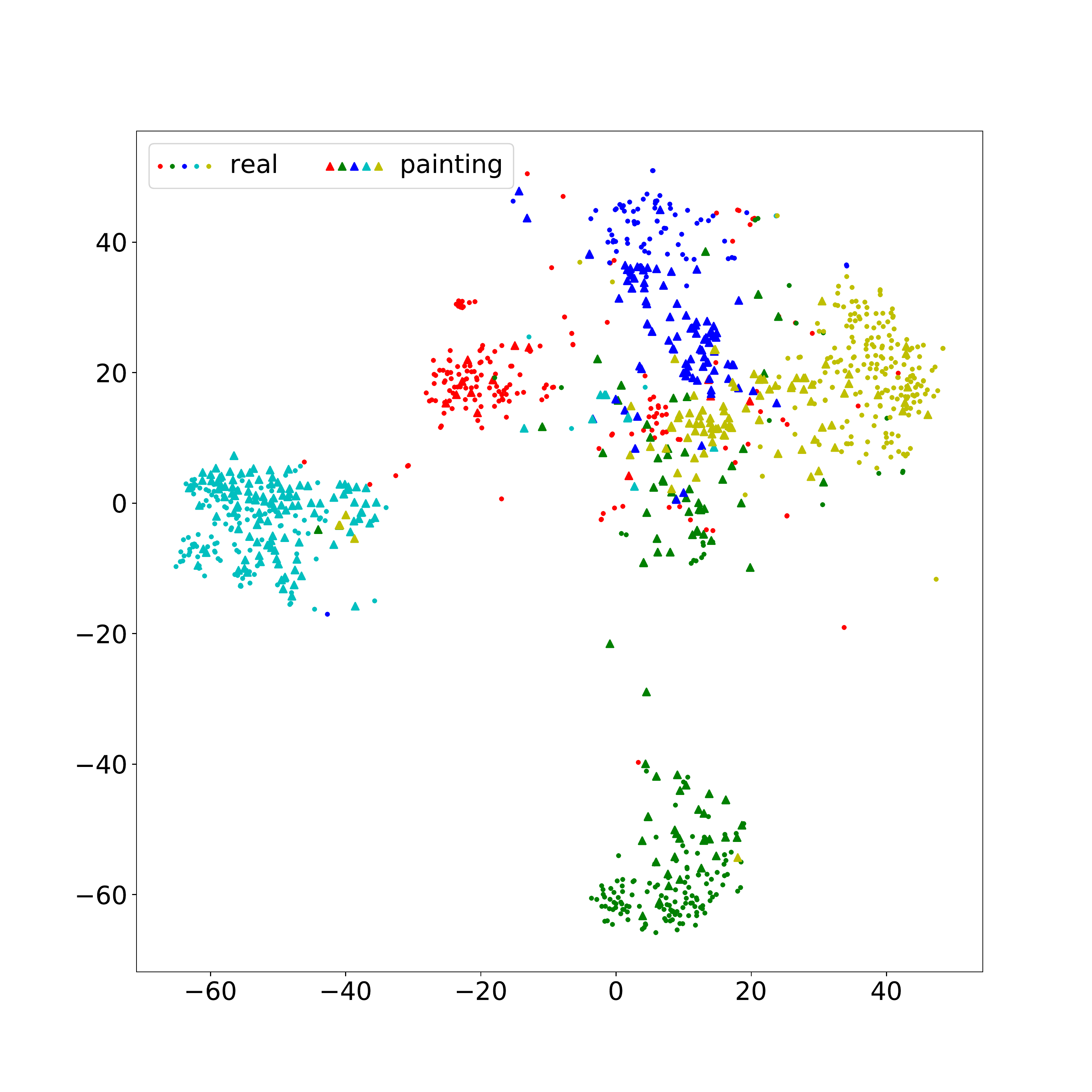}
}
\subfloat[rel vs. skt, L12]{
\includegraphics[width=0.33\linewidth]{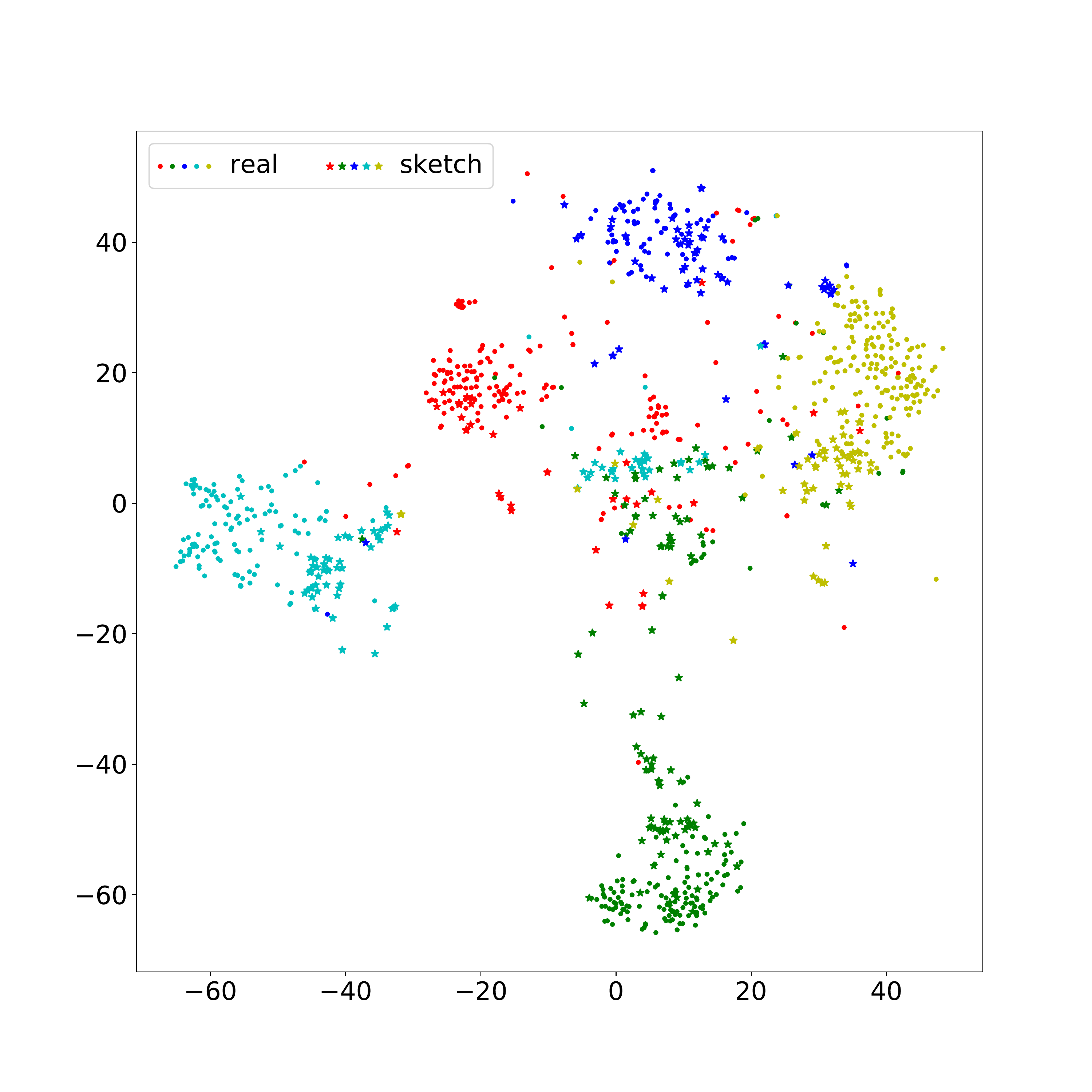}
}
\subfloat[rel vs. qcd, L12]{
\includegraphics[width=0.33\linewidth]{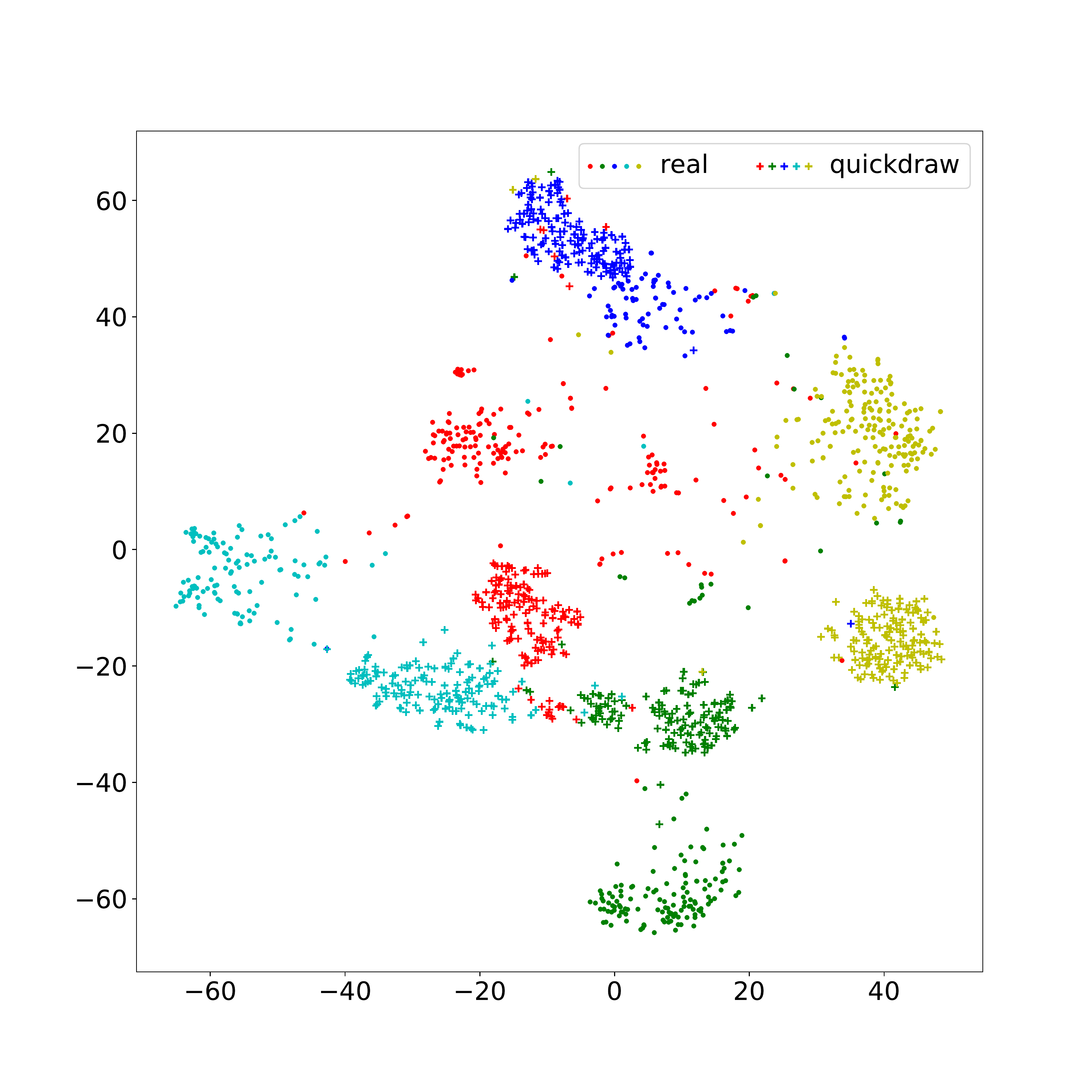}
}
\vspace{-8pt}
\caption{\textbf{T-SNE visualization results.} (a)-(c) and (d)-(e) repectively illustrates the comparison of visualizing Class Token data of \emph{real vs. painting}, \emph{real vs. sketch} and \emph{real vs. quickdraw} of layer 8 and layer 12  from four domains. Please zoom for better view.}
\label{fig: tsne_vis}
\vspace{-15pt}
\end{figure}

\subsection{Style Shifts Generalization Analysis}
%

\noindent\textbf{- ViTs have diverse performance on IID/OOD generalization gap under Style shifts.} The results on ImageNet-R are shown in \cref{fig: TX_STY} (c). As ImageNet-R only contains 200 classes of ImageNet, we follow \cite{hendrycks2020many} to record accuracy on the ImageNet subset (ImageNet-200) and regard it as the IID result. When focusing on the accuracy on ImageNet-R, we observe most ViTs beat BiTs in OOD accuracy, while having similar performance in IID/OOD generalization gap. Accordingly, ViTs do not have competitive edges in generalizing from real images to art renditions. For DomainNet, we mainly compare the models with the same scale, i.e. DeiT-S/16 and BiTs, whose results are shown in \cref{fig: domainnet}. We observe DeiT-S/16 performs better on the small-scale datasets under IID and thus the model easily outperforms BiTs in OOD accuracy. When inspecting the IID/OOD generalization gap, the results differ a lot. When models are trained on \emph{clipart} and \emph{painting}, there is no obvious difference between DeiT-S/16 and BiTs. But for \emph{real}, DeiT-S/16 leads BiTs over 4\%, which can be explained as ViTs utilize the knowledge from pre-train data better if the pre-train data and downstream data are from similar distributions.


\noindent\textbf{- ViTs shows stronger bias towards object structure.} We further investigate how models shall behave as the other available visual cues come to degrade until there only remains structural information. We illustrate examples of class \emph{parachute} of four domains and the Grad-CAM \cite{selvaraju2017grad} attention maps of both BiT and DeiT-S in \cref{fig: diff_domain_demo} (a). We shall observe that, as the color, texture, and shape cues become less and less informative from \emph{real} to \emph{quickdraw} and even there is only abstract structure preserved in \emph{quickdraw}, DeiT-S constantly concentrates on the key structural information of parachutes while BiT fails to capture such essential feature. In addition, we test the accuracies of models trained with \emph{real} on different domains. Since there are a considerable number of unrecognizable data in \emph{quickdraw}, we exclude classes on which both ViTs and CNNs achieve accuracies less than 10\%. We show the results in \cref{fig: diff_domain_demo} (b), from which we can see that the gap between ViTs and CNNs are getting larger when the tested domain contain less visual cues (\ie from \emph{real} to \emph{quickdraw}). Based on observations and analyses above, we can conclude that ViTs are less effected by the shift of color, texture, and shape features, indicating that ViTs focus more on structures.

\begin{figure*}
\centering
\subfloat[T-ADV]{
\includegraphics[width=0.23\linewidth]{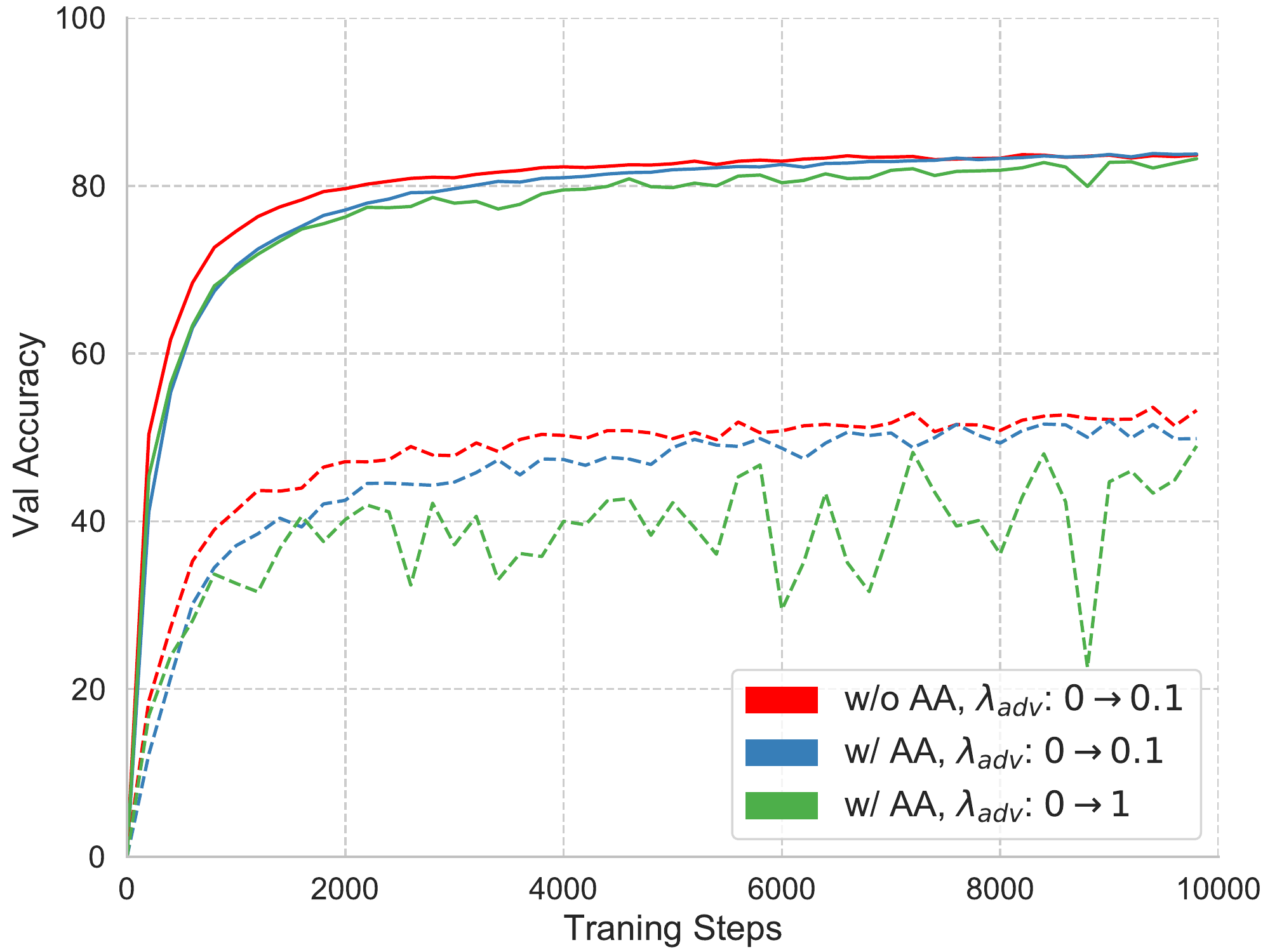}
}
\hspace{0.02\linewidth}
\subfloat[T-MME]{
\includegraphics[width=0.23\linewidth]{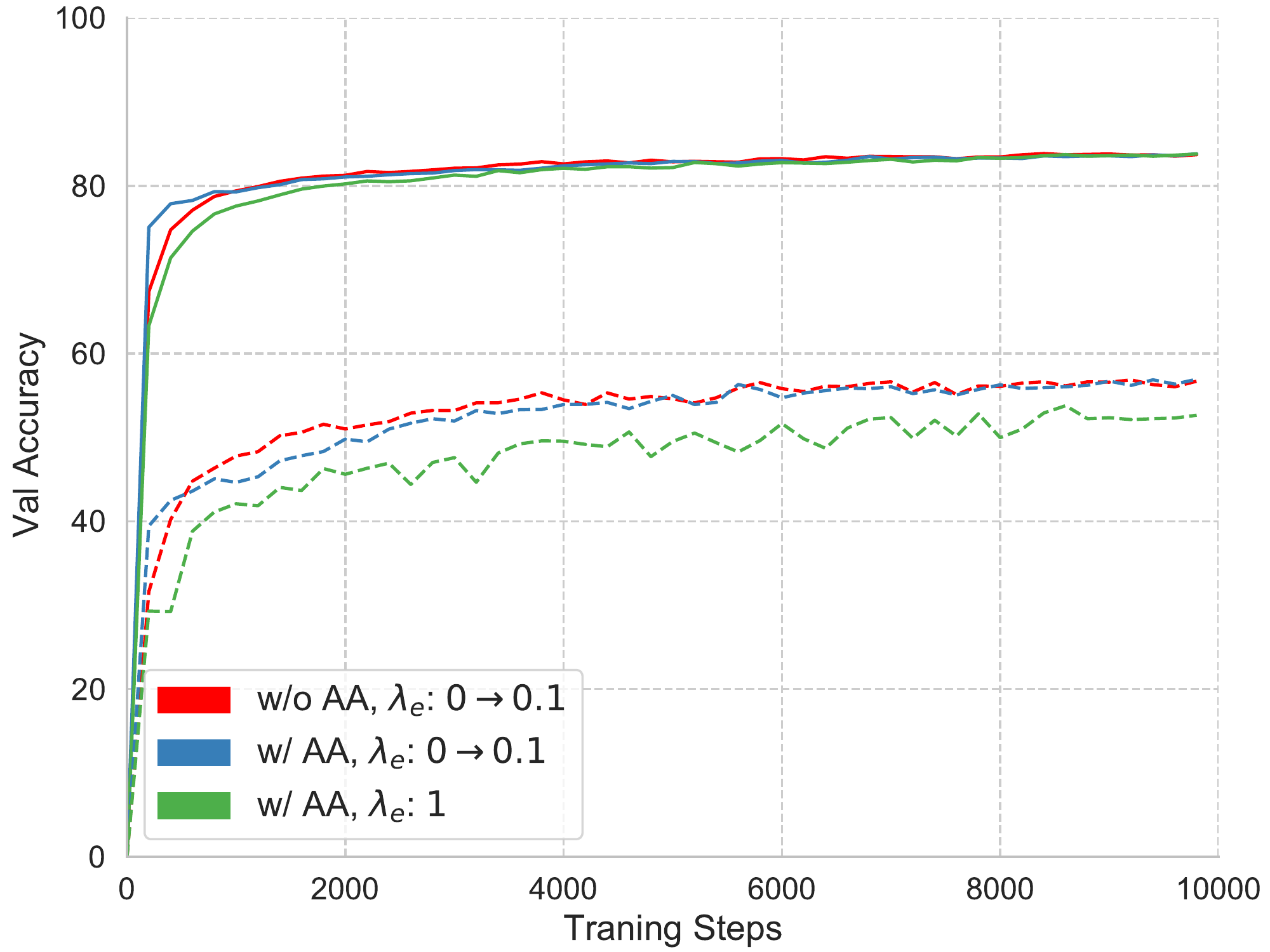}
}
\hspace{0.02\linewidth}
\subfloat[T-SSL]{
\includegraphics[width=0.23\linewidth]{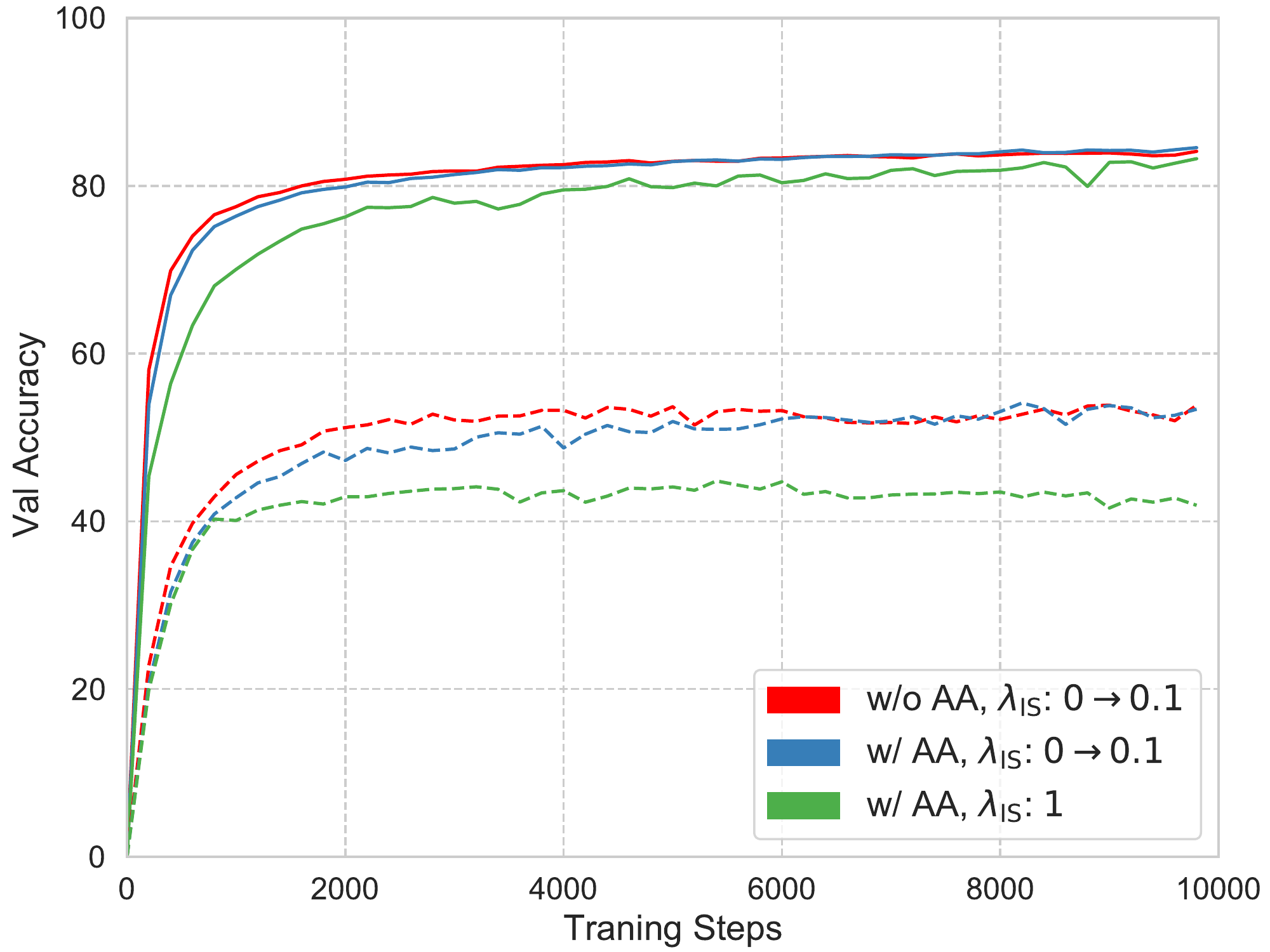}
}
\vspace{-10pt}
\caption{\textbf{Investigation of Generalization-enhanced methods with different training strategies.} (a)-(c) show training curves on both source domain and target domain. From the results, we can conclude that classical training strategies (the green lines) on CNNs are not suitable for ViTs, which need smoother strategies (the red lines) to align features in both domains.}
\label{fig:acc}
\vspace{-10pt}
\end{figure*}

\noindent\textbf{- ViTs will eliminate different levels of DS in different layers.} 
We select a set of classes of four domains in DomainNet shown in \cref{fig: diff_domain_demo} (a) and the class lists are shown in the supplementary material. By extracting the intermediate Class Token and implementing dimensionality reduction via T-SNE technique \cite{van2008visualizing}, we generate the visualizations of Class Token data of layer 8 and layer 12 from four domains and respectively show the comparison of \emph{real vs. painting}, \emph{real vs. sketch} and \emph{real vs. quickdraw}. As shown in \cref{fig: tsne_vis}, we can first observe from pictures in the first row that data from different domains are clustered together to a certain extent only in the \emph{real vs. painting} condition at layer 8. As for \emph{real vs. sketch}, the data become well clustered until at layer 12 (\cref{fig: tsne_vis} (e)), whereas the \emph{real vs. quickdraw} condition fails to mix up data from different domains together but there exists the decision boundary that can well divide data of different classes for both domains at layer 12 (\cref{fig: tsne_vis} (f)). From the above analysis, we conclude that ViTs will eliminate different levels of DS in different layers.



\section{Studies on Generalization-Enhanced ViTs} \label{sec: enhance}



\noindent\textbf{Settings.}
We use DomainNet \cite{peng2019moment} for the following experiments. Following \cite{saito2019semi}, we focus on the 7 scenarios listed in \cref{table:DA results}. To make a full comparison, we implement these enhancing techniques on two representative CNNs VGG-16 and BiT, and two ViTs including DeiT-S/16 and DeiT-B/16. We explore their performance on both the vanilla version and the generalization-enhanced version. 
Implementation details can be found in supplementary materials.

\begin{table}[t]
  \caption{\textbf{Results of Generalization-enhanced methods.} Specifically, we compare three types of GE-ViTs with their corresponding CNNs. From the results we could conclude that 1) equipped with GE-ViTs, we achieve significant performance boosts towards out-of-distribution data by 4\% from vanilla ViTs. 2) three GE-ViTs have almost the same improvement from vanilla models on OOD accuracy. 3) for the enhanced transformer models, larger ViTs still benefit more for the out-of-distribution generalization.}
  \label{table:DA results}
  \centering
  \resizebox{1.0\linewidth}{!}{
  \begin{tabular}{cccccccccc}
    \toprule
    Model & Method & R to C & R to P & P to C & C to S & S to P & R to S & P to R & Avg. \\ \midrule
    \multirow{4}{*}{DeiT-B/16} & - & 54.64 & 48.40 & 40.37 & 45.69 & 36.75 & 41.31 & 55.33 & 46.07\\
    ~ & T-ADV & 58.19 & 50.85 & 41.91 & 51.18 & 46.12 & 47.47 & 55.65 & 50.20\\
    ~ & T-MME & \textbf{60.59} & \textbf{51.98} & 42.30 & 50.32 & 45.79 & \textbf{47.92} & 54.87 & 50.54\\ 
    ~ & T-SSL & 56.80 & 49.06 & \textbf{45.96} & \textbf{51.79} & \textbf{46.95} & 45.95 & \textbf{60.98} & \textbf{51.07} \\ \midrule
    \multirow{4}{*}{DeiT-S/16} & - & 50.60 & 45.82 & 36.09 & 43.39 & 35.24 & 39.29 & 52.08 & 43.22\\
    ~ & T-ADV & 53.60 & 47.84 & 37.99 & 47.10 & 41.61 & 41.94 & 52.82 & 46.13\\
    ~ & T-MME & \textbf{56.86} & \textbf{49.15} & 38.97 & 46.48 & 42.95 & \textbf{42.07} & 52.49 & 47.00\\ 
    ~ & T-SSL & 53.86 & 46.71 & \textbf{42.79} & \textbf{47.25} & \textbf{43.01} & 40.94 & \textbf{57.07} & \textbf{47.37} \\ \midrule
    \multirow{4}{*}{BiT} & - & 42.18 & 41.14 & 30.72 & 37.01 & 28.23 & 32.64 & 48.54 & 36.78 \\
    ~ & DANN \cite{ganin2015unsupervised} & 45.20 & 42.86 & 32.96 & 40.44 & 36.63 & 35.26 & 49.25 & 40.37\\
    ~ & MME \cite{saito2019semi} & 50.21 & \textbf{44.61} & 34.75 & 40.27 & 38.41 & 37.83 & 47.58 & 41.95\\ 
    ~ & SSL \cite{yue2021prototypical} & \textbf{52.55} & 42.80 & \textbf{39.03} & \textbf{45.72} & \textbf{39.08} & \textbf{39.65} & \textbf{56.07} & \textbf{44.98} \\ \midrule
    \multirow{4}{*}{VGG-16} & - & 39.39 & 37.32 & 26.36 & 32.96 & 25.55 & 27.79 & 45.70 & 33.58\\
    ~ & DANN \cite{ganin2015unsupervised} & 43.26 & 40.09 & 28.68 & 36.22 & 31.63 & \textbf{35.45} & 44.73 & 37.15 \\
    ~ & MME \cite{saito2019semi} & 42.65 & \textbf{42.46} & 27.41 & \textbf{36.93} & 33.94 & 32.58 & \textbf{45.87} & 37.41 \\ 
    ~ & SSL \cite{yue2021prototypical} & \textbf{43.79} & 41.88 & \textbf{32.19} & 35.73 & \textbf{36.99} & 31.05 & 55.18 & \textbf{39.54} \\

    \bottomrule
  \end{tabular}
  }
  \vspace{-10pt}
\end{table}

\noindent\textbf{Performance Analysis.}
The results of three GE-ViTs comparing with CNNs are shown in \cref{table:DA results}. From the results we have the following observations: \textbf{1)} equipped with GE-ViTs, we achieve significant performance boosts towards out-of-distribution data by 4\% from vanilla ViTs. \textbf{2)} Three GE-ViTs have almost the same improvement from vanilla models on OOD accuracy. In contrast, CNNs benefit more from the self-supervised learning method than the others. \textbf{3)} DeiT-B/16 has a larger gain on those enhancing methods than DeiT-S/16. Therefore, we conclude that \textbf{1)} ViTs and CNNs share many characteristics, and both can be beneficial from the generalization-enhancement methods. \textbf{2)} For the enhanced transformer models, larger ViTs still benefit more for the out-of-distribution generalization.






\noindent\textbf{Smooth Feature Alignment.}
\cref{fig:acc} shows the performance of GE-ViTs with different training strategies. The green line represents the same training strategies used in CNNs. The other two lines use smoother strategies. From the comparison of these strategies, we observe that \textbf{1)} the generally used automated augmentation schemes shall cause performance degradation on T-ADV while they have little influence on T-MME and T-SSL. \textbf{2)} smoother learning strategies are significant for ViT convergence, especially in the adversarial training mode. As for T-MME and T-SSL, smoothness of auxiliary losses also significantly improves the performance. Based on these observations, we conclude that GE-ViTs are more sensitive to the hyper-parameters than their corresponding CNN models.

\section{Discussion and Conclusion}
We provide a comprehensive study on the OOD generalization of ViTs, with the following contributions:
\textbf{1)} We define a taxonomy on data distribution shifts according to the modified semantic concepts in images.
\textbf{2)} We perform a comprehensive study on OOD generalization and inductive bias properties of ViTs under the five categorized distribution shifts. Several valuable observations are obtained.
\textbf{3)} We further improve the OOD generalization of ViTs by designing GE-ViTs through adversarial learning, information theory, and self-supervised learning with smoother training strategies.
Our work serves as an early attempt, thus there is plenty of room for developing more powerful GE-ViTs.

\noindent\textbf{Broader Impacts.} Some models utilized in this paper demand a large number of computing resources for training procedures. The consumption of electricity may exert an environmental impact. 
\section*{Acknowledgements}
This work is supported by NTU NAP, and under the RIE2020 Industry Alignment Fund – Industry Collaboration Projects (IAF-ICP) Funding Initiative, as well as cash and in-kind contribution from the industry partner(s).

{\small
\bibliographystyle{ieee_fullname}
\bibliography{ref}
}

\clearpage
\appendix

\begin{figure*}[t]
\centering
\includegraphics[width=0.98\linewidth]{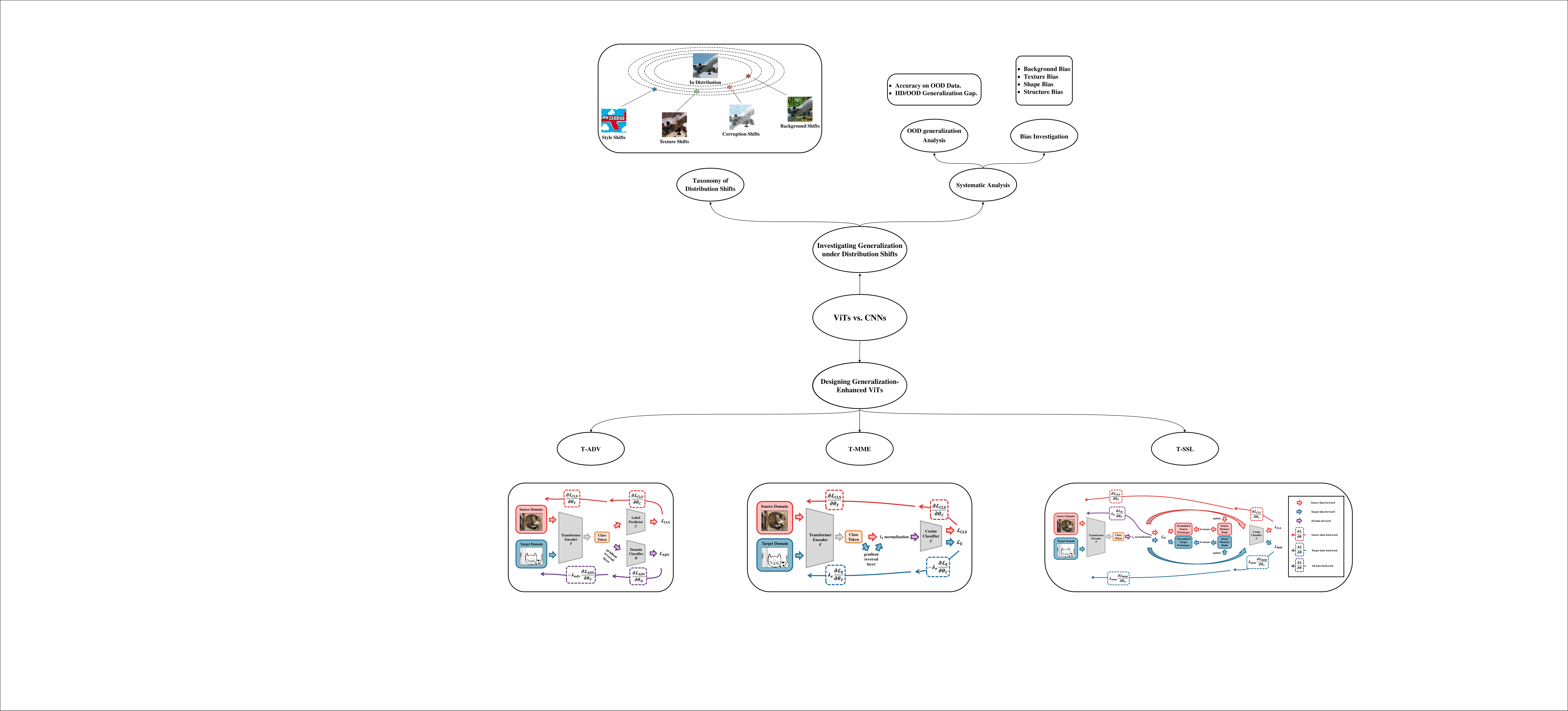}
\caption{\textbf{A brief outline of our work.}}
\label{fig:gist}
\end{figure*}

\section{Brief Outline of Our Work}
To help reviewers better understand our work, we make an outline of this paper in \cref{fig:gist}. In brief, we have investigated generalization of ViTs under distribution shifts and designed generalization enhanced ViTs.

\section{Experimental Setup}

\subsection{Dataset Zoo}

\begin{itemize}[leftmargin=0.55cm]

\item \textbf{ImageNet-9} \cite{xiao2020noise} is adopted for background shifts. ImageNet-9 is a variety of 9-class datasets with different foreground-background recombination plans, which helps disentangle the impacts of foreground and background signals on classification. In our case, we use the four varieties of generated background with foreground unchanged, including 'Only-FG', 'Mixed-Same', 'Mixed-Rand' and 'Mixed-Next'. The 'Original' data set is used to represent in-distribution data. 

\item \textbf{ImageNet-C} \cite{hendrycks2018benchmarking} is used to examine generalization ability under corruption shifts. ImageNet-C includes 15 types of algorithmically generated corruptions, grouped into 4 categories: ‘noise’, ‘blur’, ‘weather’, and ‘digital’. Each corruption type has five levels of severity, resulting in 75 distinct corruptions. 

\item \textbf{Cue Conflict Stimuli} and \textbf{Stylized-ImageNet} are used to investigate generalization under texture shifts. Utilizing style transfer, \cite{geirhos2018imagenettrained} generated \textbf{Cue Conflict Stimuli} benchmark with conflicting shape and texture information, that is, the image texture is replaced by another class with other object semantics preserved. In this case, we respectively report the shape and texture accuracy of classifiers for analysis. Meanwhile, \textbf{Stylized-ImageNet} is also produced in \cite{geirhos2018imagenettrained} by replacing textures with the style of randomly selected paintings through AdaIN style transfer \cite{huang2017arbitrary}. 


\item \textbf{ImageNet-R} \cite{hendrycks2020many} and \textbf{DomainNet} \cite{peng2019moment} are used for the case of style shifts. ImageNet-R \cite{hendrycks2020many} contains 30000 images with various artistic renditions of 200 classes of the original ImageNet validation data set. The renditions in ImageNet-R are real-world, naturally occurring variations, such as paintings or embroidery, with textures and local image statistics which differ from those of ImageNet images. DomainNet \cite{peng2019moment} is a recent benchmark dataset for large-scale domain adaptation that consists of 345 classes and 6 domains. As labels of some domains are very noisy, we follow the 7 distribution shift scenarios in \cite{saito2019semi} with 4 domains (Real, Clipart, Painting, Sketch) picked.

\end{itemize}

\begin{figure*}[t]
\centering
\subfloat[bowtie]{
\includegraphics[width=0.48\linewidth]{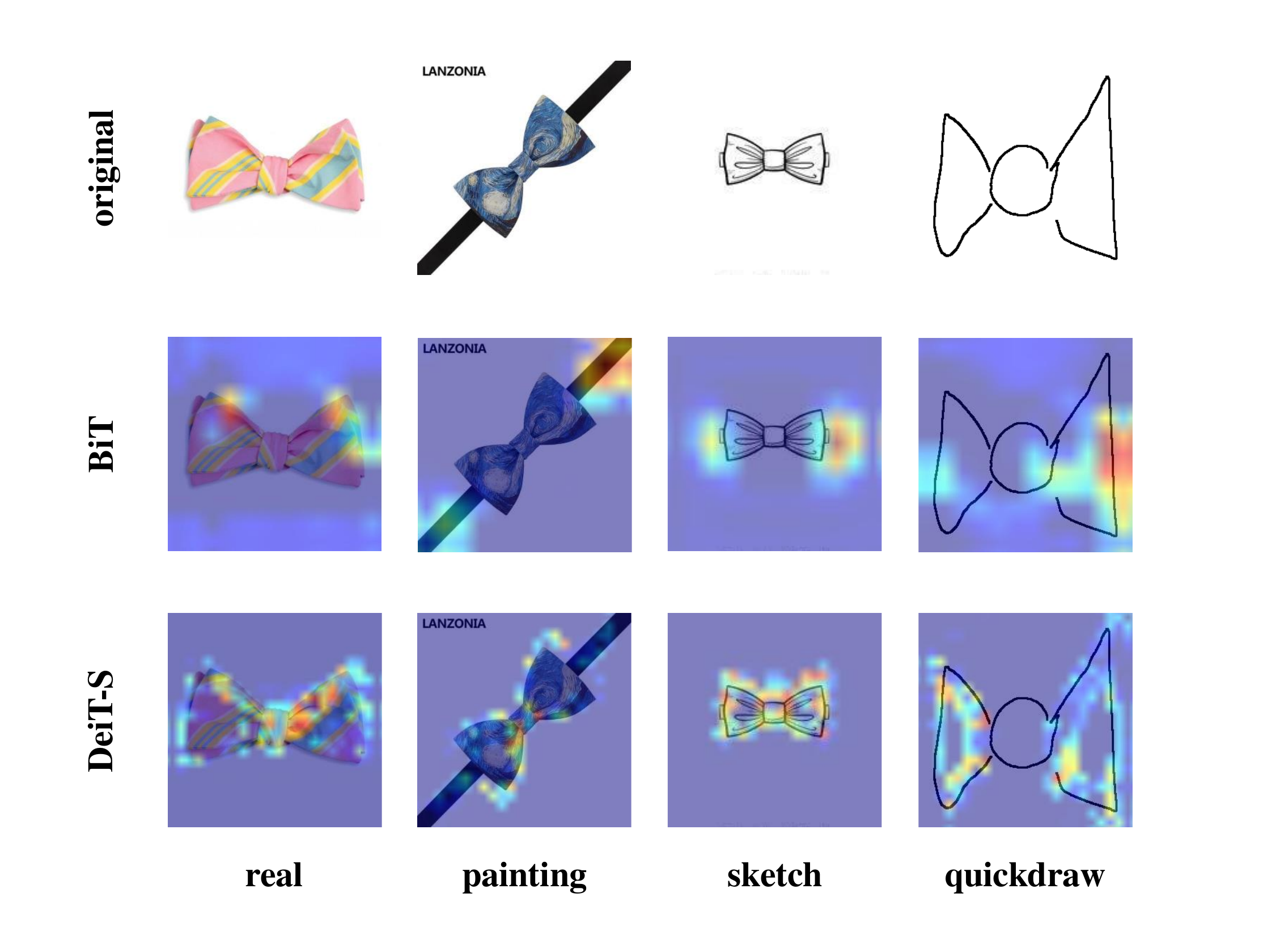}
}
\subfloat[coffee cup]{
\includegraphics[width=0.48\linewidth]{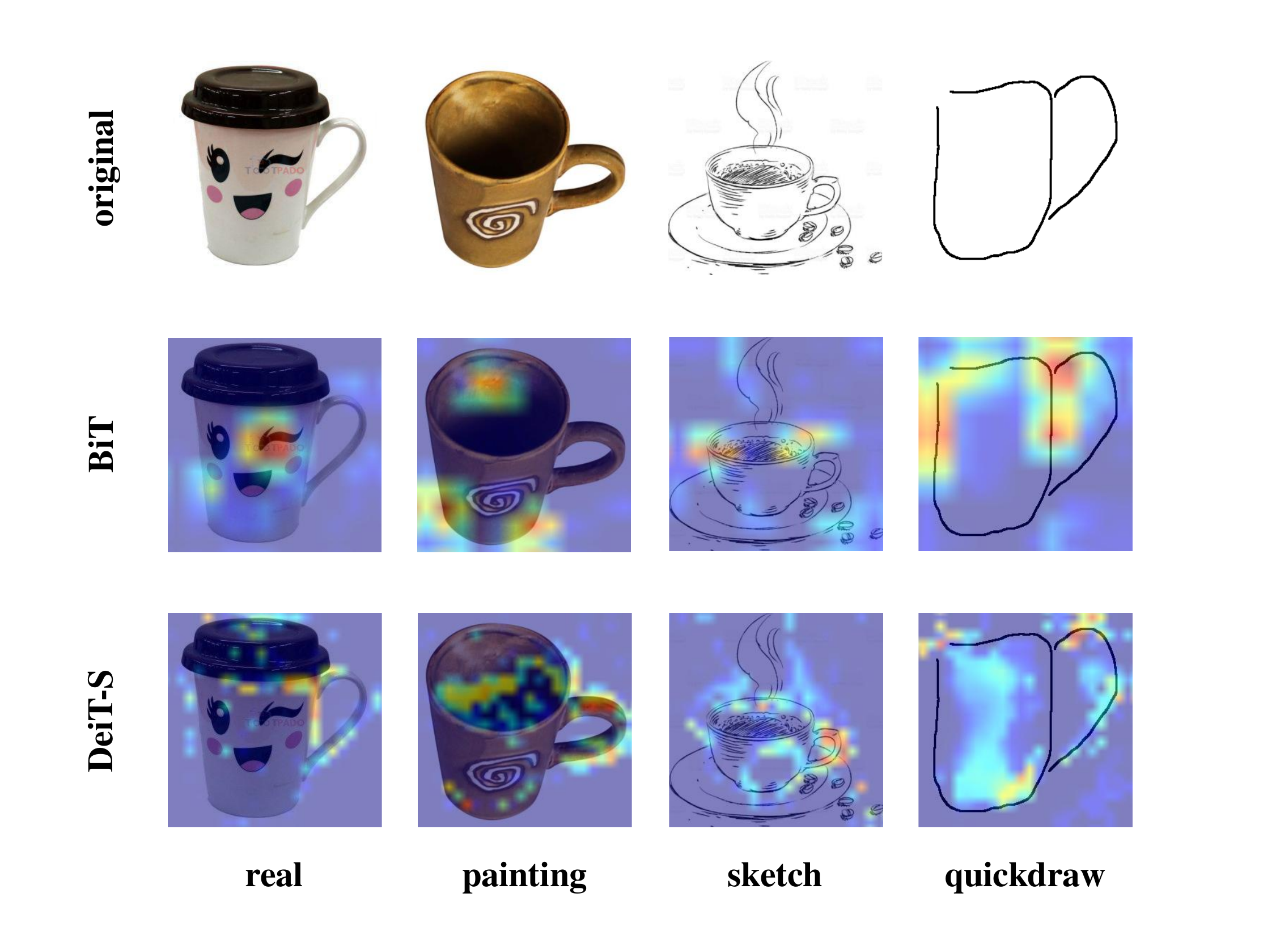}
}\\
\subfloat[duck]{
\includegraphics[width=0.48\linewidth]{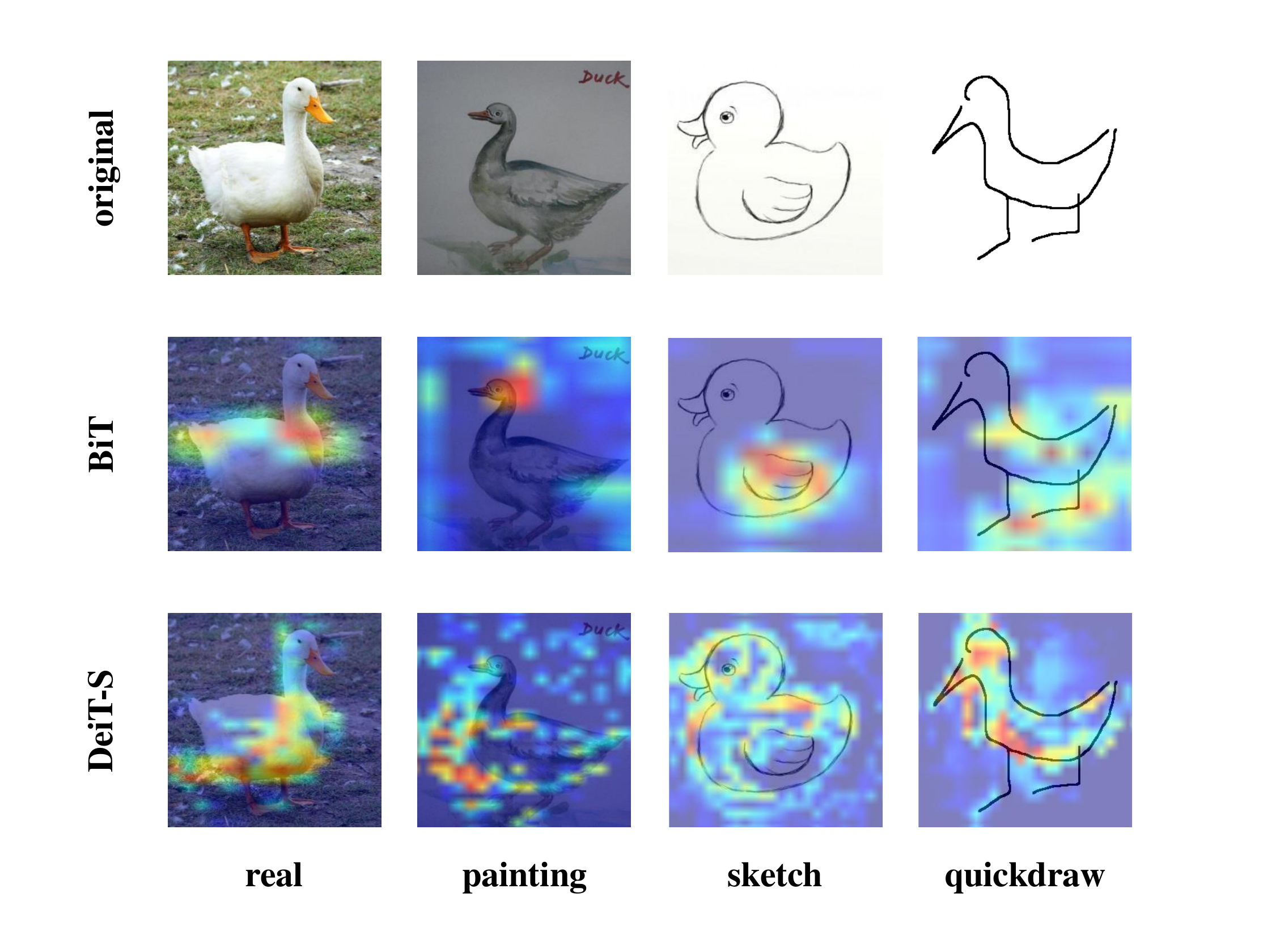}
}
\subfloat[hourglass]{
\includegraphics[width=0.48\linewidth]{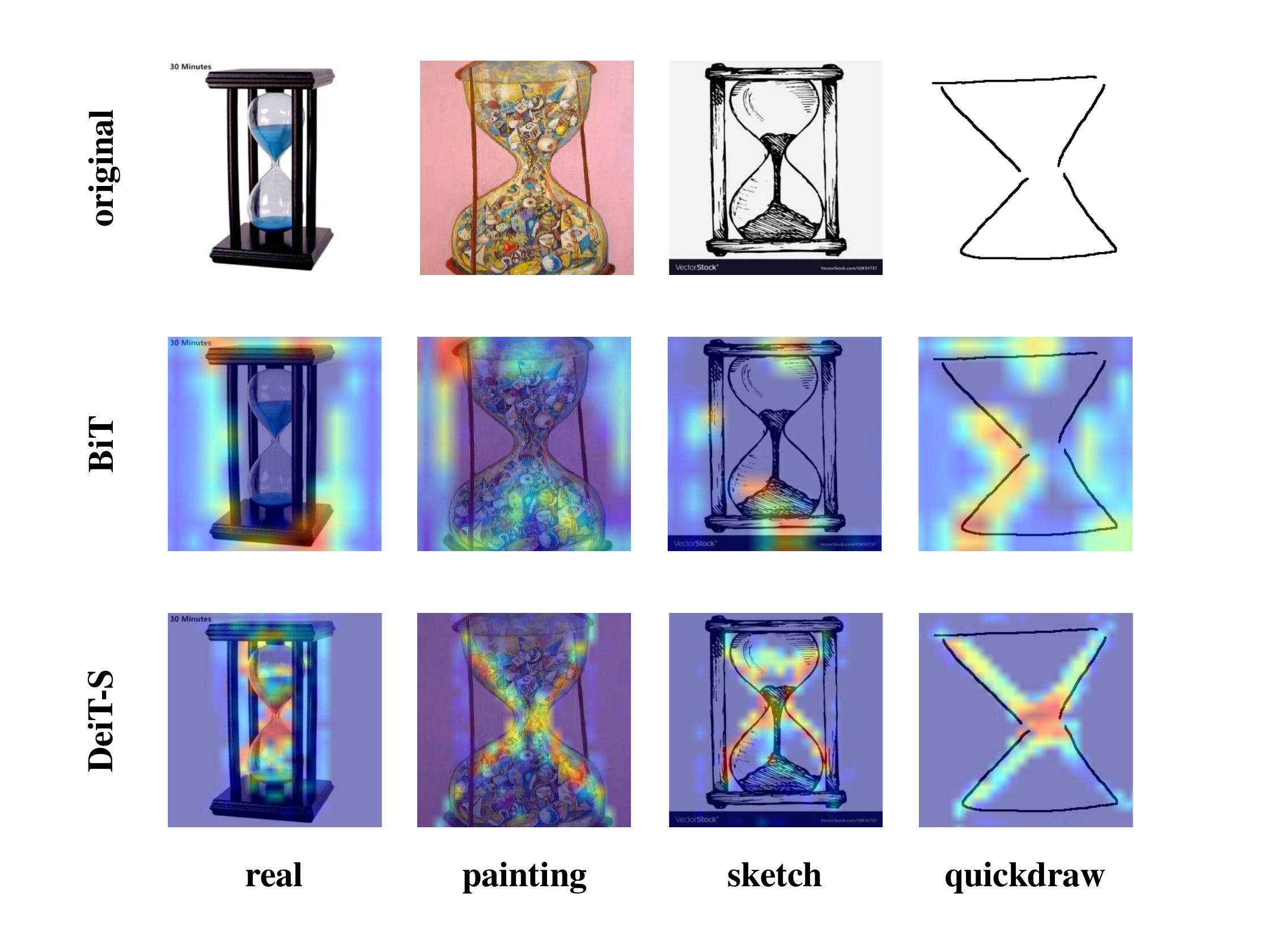}
}
\caption{\textbf{Grad-CAM visualization on more classes.} (a)-(d) corresponds to the class of \emph{bowtie}, \emph{coffee cup}, \emph{duck} and \emph{hourglass} respectively. All attention maps are generated using models trained on \emph{real}.}
\label{fig:more Grad-CAM}
\vspace{-10pt}
\end{figure*}

\begin{table}
   \caption{\textbf{Basic description of used model architectures.}}
   \label{table:arch}
   \centering
   \resizebox{1.0\linewidth}{!}{
   \begin{tabular}{cccccc}
     \toprule
     Model & patch size & embedding dimension & \#heads & \#layers & \#params \\ \midrule
     DeiT-Ti/16 & 16$\times$16 & 192 & 3 & 12 & 5M \\
     DeiT-S/16 & 16$\times$16 & 384 & 6 & 12 & 22M \\
     DeiT-B/16 & 16$\times$16 & 768 & 12 & 12 & 86M \\
     DeiT-B/32 & 32$\times$32 & 768 & 12 & 12 & 86M \\
     DeiT-L/16 & 16$\times$16 & 1024 & 16 & 24 & 307M \\
     BiT-S-R50X1 & - & - & - & - & 23M \\
    
     \bottomrule
   \end{tabular}
   }
\end{table}

\subsection{Implementation Details} \label{sec:implement}

\subsubsection{Vanilla Model Implementation}

\begin{itemize}[leftmargin=0.55cm]

\item \textbf{DeiT.} For Vision Transformers, we pre-train the DeiT models on the ImageNet dataset with the AdamW optimizer \cite{loshchilov2018fixing}, a batch size of 1024 and the resolution of $224 \times 224$. The learning rate is linearly ramped up during the first 5 epochs to its base value determined with the following linear scaling rule: $lr$ = 0.0005*batch size/512. After the warmup, we decay the learning rate with a cosine schedule \cite{loshchilov2016sgdr} with the weight decay = 0.05 and train for 300 epochs. We follow the data augmentations scheme in the official paper \cite{touvron2020training}. For the downstream fine-tuning, we scale up the resolution to $384 \times 384$ by adopting perform 2D interpolation of the pre-trained position embeddings proposed in \cite{dosovitskiy2020image}. We train models with learning rate $lr$ = 5e-6, weight decay = 1e-8 for 75000 iters and the other settings identical to pre-training stage.

\item \textbf{BiT.} BiT models use the augmentation schemes mentioned in \cite{DBLP:conf/eccv/KolesnikovBZPYG20}. We train them upstream using SGD with momentum. We use an initial learning rate of 0.03, weight decay 0.0001, and momentum 0.9. We train for 90 epochs and decay the learning rate by a factor of 10 at 30, 60, and 80 epochs. We use a global batch size of 4096 and multiply the learning rate by batch size/256 with the resolution $224 \times 224$. For downstream fine-tuning, we use SGD with an initial learning rate of 0.003, momentum 0.9, and batch size 512 with the resolution $384 \times 384$.

\item \textbf{BiT$_{da}$.} BiT$_{da}$ models use identical data augmentation strategy from DeiTs. The other training setups are consistent with BiT models.

\end{itemize}

The basic description of used model architectures is shown in \cref{table:arch}.

\subsubsection{Generalization-Enhanced Model Implementation}

\begin{itemize}[leftmargin=0.55cm]

\item \textbf{T-ADV.} T-ADV consists of a feature encoder, a label predictor, and a domain classifier. The feature encoder is implemented by the DeiT backbone and the label predictor is a linear layer projecting CLS token to logit. The domain classifier is a three-layer MLP with hidden dimension 1024, aiming at predicting domain labels. We implement T-ADV on downstream tasks using ImageNet pre-trained DeiT backbones. We train models with learning rate $lr$ = 5e-5 and weight decay = 1e-7 for 10000 iters except that the domain classifier use a learning rate $lr$ = 2.5e-4. Furthermore, we make some adjustments to the training scheme due to the special architectures of Vision Transformers. Based on our practice, we restrict the magnitude of gradient reverse coefficient $\lambda_{adv}$ by further multiplying 0.1 from the original setting, to avoid great fluctuation during training. These adjustments contribute to a stable adversarial training process.

\item \textbf{T-MME.} T-MME consists of a feature encoder and a cosine similarity-based classifier. The classifier is implemented by a three-layer MLP with a hidden size of 1024. The learning rate of this part is scaled up 10 times, which is consistent with the original setting. We make further adjustments on DeiT architectures by introducing the adaptive update scheme in \cite{ganin2015unsupervised} on the coefficient $\lambda_{e}$ from 0 to 0.1, instead of the original constant setting. The other training setups are consistent with T-ADV.

\begin{table*}[t]
  \caption{\textbf{SSL ablation.}}
  \label{table:SSL ablation}
  \centering
  \small
  \begin{tabular}{cccccccccc}
    \toprule
    Model & Method & R to C & R to P & P to C & C to S & S to P & R to S & P to R & Avg. \\ \midrule
    \multirow{4}{*}{DeiT-S/16} & - & 50.60 & 45.82 & 36.09 & 43.39 & 35.24 & 39.29 & 52.08 & 43.22\\
    ~ & T-MIM & 53.67 & 44.80 & 42.31 & 47.00 & \textbf{43.28} & \textbf{41.70} & \textbf{58.38} & 47.30\\ 
    ~ & T-IS & 53.31 & \textbf{47.19} & 40.93 & 46.71 & 41.93 & 38.77 & 55.62 & 46.35\\
    ~ & T-SSL & \textbf{53.86} & 46.71 & \textbf{42.79} & \textbf{47.25} & 43.01 & 40.94 & 57.07 & \textbf{47.37} \\ \midrule
    \multirow{4}{*}{BiT} & - & 42.18 & 41.14 & 30.72 & 37.01 & 28.23 & 32.64 & 48.54 & 36.78 \\
    ~ & MIM \cite{yue2021prototypical} & \textbf{53.02} & 41.64 & \textbf{40.24} & 45.10 & 38.26 & \textbf{40.17} & 54.70 & 44.73 \\
    ~ & IS \cite{yue2021prototypical} & 48.31 & \textbf{44.12} & 36.32 & 43.84 & 38.30 & 35.81 & 53.31 & 42.85 \\
    ~ & SSL \cite{yue2021prototypical} & 52.55 & 42.80 & 39.03 & \textbf{45.72} & \textbf{39.08} & 39.65 & \textbf{56.07} & \textbf{44.98} \\ \midrule
    \multirow{4}{*}{VGG-16} & - & 39.39 & 37.32 & 26.36 & 32.96 & 25.55 & 27.79 & 45.70 & 33.58\\
    ~ & MIM \cite{yue2021prototypical} & \textbf{48.41} & 42.18 & \textbf{36.34} & \textbf{43.08} & \textbf{38.45} & \textbf{37.51} & 54.32 & \textbf{42.89} \\
    ~ & IS \cite{yue2021prototypical} & 42.05 & \textbf{42.36} & 31.30 & 38.68 & 36.59 & 30.74 & 51.07 & 38.97 \\
    ~ & SSL \cite{yue2021prototypical} & 43.79 & 41.88 & 32.19 & 35.73 & 36.99 & 31.05 & \textbf{55.18} & 39.54 \\
    \bottomrule
  \end{tabular}
  
\end{table*}

\begin{table*}[t]
  \caption{\textbf{Results of Generalization-enhanced methods under corruption shifts including noises and blurs.}}
  \label{table:additional DA results1}
  \centering
  \small
  \begin{tabular}{ccccccccccc}
    \toprule
    \multirow{2}{*}{Model} & \multirow{2}{*}{Method} & \multicolumn{4}{c}{Noise} & \multicolumn{5}{c}{Blur} \\ \cmidrule(lr){3-6} \cmidrule(lr){7-11}
    ~ & ~ & Gauss. & Impulse & Shot & Avg. & Defocus & Glass & Motion & Zoom & Avg. \\ \midrule
    \multirow{6}{*}{DeiT-S/16} & - & 55.51 & 54.70 & 54.74 & 54.98 & 45.11 & 27.82 & 51.51 & 41.51 & 41.49 \\
    ~ & T-ADV & 58.96 & 62.35 & 60.02 & 60.44 & \textbf{55.95} & 54.94 & 62.38 & 62.07 & 58.84 \\
    ~ & T-MME & \textbf{60.40} & \textbf{63.85} & \textbf{60.96} & \textbf{61.74} & 55.86 & \textbf{56.02} & \textbf{63.69} & \textbf{63.35} & \textbf{59.73}\\ 
    ~ & T-MIM & 59.10 & 62.17 & 59.37 & 60.21 & 53.97 & 54.39 & 62.00 & 61.46 & 57.96\\
    ~ & T-IS & 39.78 & 41.18 & 39.67 & 40.21 & 32.87 & 21.48 & 41.59 & 34.16 & 32.52\\
    ~ & T-SSL & 56.54 & 60.26 & 56.99 & 57.93 & 50.24 & 51.56 & 59.68 & 59.04 & 55.13\\ \midrule
    \multirow{6}{*}{BiT} & - & 37.48 & 33.46 & 34.70 & 35.21 & 22.62 & 10.04 & 30.54 & 31.44 & 23.66\\
    ~ & DANN \cite{ganin2015unsupervised} & 37.21 & 41.08 & 39.85 & 39.38 & \textbf{30.55} & 28.54 & 42.70 & 40.60 & \textbf{35.60}\\
    ~ & MME \cite{saito2019semi} & \textbf{43.51} & \textbf{46.14} & \textbf{43.81} & \textbf{44.49} & 15.07 & \textbf{30.75} & \textbf{47.17} & \textbf{42.82} & 33.95\\ 
    ~ & MIM \cite{yue2021prototypical} & 36.23 & 36.33 & 35.81 & 36.12 & 23.95 & 25.59 & 40.11 & 34.68 & 31.08\\ 
    ~ & IS \cite{yue2021prototypical} & 18.33 & 17.24 & 17.04 & 17.54 & 9.77 & 6.07 & 15.41 & 17.33 & 12.15\\ 
    ~ & SSL \cite{yue2021prototypical} & 35.00 & 35.28 & 34.95 & 35.08 & 25.90 & 26.19 & 36.42 & 20.15 & 27.16\\ 

    \bottomrule
  \end{tabular}
  
\end{table*}

\begin{table*}[t]
  \caption{\textbf{Results of Generalization-enhanced methods under corruption shifts including weathers and digital corruptions.}}
  \label{table:additional DA results2}
  \centering
  \small
  \begin{tabular}{cccccccccccc}
    \toprule
    \multirow{2}{*}{Model} & \multirow{2}{*}{Method} & \multicolumn{5}{c}{Weather} & \multicolumn{5}{c}{Digital} \\ \cmidrule(lr){3-7} \cmidrule(lr){8-12}
    ~ & ~ & Bright & Fog & Frost & Snow & Avg. & Contrast & Elastic & JPEG & Pixel & Avg. \\ \midrule
    \multirow{6}{*}{DeiT-S/16} & - & 72.11 & 53.67 & 50.45 & 53.79 & 57.51 & 66.58 & 63.10 & 61.25 & 60.51 & 62.86\\
    ~ & T-ADV & 71.54 & 67.07 & 59.81 & 66.43 & 66.21 & 67.73 & 69.83 & 64.43 & 68.25 & 67.56\\
    ~ & T-MME & 71.93 & \textbf{67.62} & \textbf{60.57} & \textbf{67.03} & \textbf{66.79} & \textbf{68.44} & 70.04 & \textbf{65.65} & \textbf{68.92} & \textbf{68.26}\\ 
    ~ & T-MIM & \textbf{72.37} & 66.98 & 59.76 & 66.38 & 66.37 & 68.22 & \textbf{70.19} & 64.39 & 68.56 & 67.84\\
    ~ & T-IS & 66.50 & 51.91 & 39.62 & 51.65 & 52.42 & 56.06 & 58.77 & 53.53 & 50.07 & 54.60\\
    ~ & T-SSL & 70.85 & 65.42 &57.04 & 64.80 & 64.53 & 66.38 &68.83 & 63.71 & 66.43 & 66.34\\ \midrule
    \multirow{6}{*}{BiT} & - & \textbf{65.96} & 49.08 & 32.23 & 34.68 & 45.49 & 54.57 & 44.61 & 52.74 & 48.27 & 50.04\\
    ~ & DANN \cite{ganin2015unsupervised} & 60.88 & 54.67 & 37.95 & 48.10 & 50.40 & 55.75 & 56.12 & 48.52 & 54.57 & 53.74\\
    ~ & MME \cite{saito2019semi} & 61.94 & \textbf{57.05} & \textbf{41.45} & \textbf{52.43} & \textbf{53.22} & \textbf{57.46} & \textbf{57.93} & \textbf{51.28} & \textbf{56.61} & \textbf{55.82}\\
    ~ & MIM \cite{yue2021prototypical} & 60.38 & 56.37 & 34.66 & 48.92 & 50.08 & 56.29 & 57.17 & 50.55 & 55.92 & 54.98\\
    ~ & IS \cite{yue2021prototypical} & 49.45 & 34.50 & 17.77 & 22.01 & 30.93 & 35.03 & 32.27 & 32.19 & 28.44 & 31.98\\
    ~ & SSL \cite{yue2021prototypical} & 54.78 & 48.88 & 34.25 & 42.20 & 45.03 & 49.26 & 49.54 & 44.67 & 15.68 & 39.79\\
    \bottomrule
  \end{tabular}
  
\end{table*}

\begin{table*}[t]
  \caption{\textbf{Results of Generalization-enhanced methods under background shifts.}}
  \label{table:additional DA results3}
  \centering
  \small
  \begin{tabular}{ccccccc}
    \toprule
    \multirow{2}{*}{Model} & \multirow{2}{*}{Method} & \multicolumn{5}{c}{Background} \\ \cmidrule(lr){3-7}
    ~ & ~ & Only-FG & Mixed-Same & Mixed-Rand & Mixed-Next & Avg.  \\ \midrule
    \multirow{6}{*}{DeiT-S/16} & - & 88.80 & 90.21 & 84.08 & 82.70 & 86.44  \\
    ~ & T-ADV & 94.26 & 96.54 & 92.79 & 92.86 & 94.11 \\
    ~ & T-MME & 93.89 & 96.39 & 92.13 & 91.69 & 93.52 \\ 
    ~ & T-MIM & \textbf{95.51} & 96.83 & 92.72 & 92.42 & 84.37  \\
    ~ & T-IS & 95.36 & 96.17 & 88.67 & 87.57 & 91.94  \\
    ~ & T-SSL & 94.92 & \textbf{96.98} & \textbf{93.01} & \textbf{93.16} & \textbf{94.52}  \\ \midrule
    \multirow{6}{*}{BiT} & - & 82.04 & 83.81 & 77.07 & 74.19 & 79.27  \\
    ~ & DANN \cite{ganin2015unsupervised} & \textbf{94.19} & \textbf{95.44} & \textbf{91.25} & \textbf{91.76} & \textbf{93.16}  \\
    ~ & MME \cite{saito2019semi} & 87.57 & 92.75 & 83.72 & 87.27 & 87.83  \\ 
    ~ & MIM \cite{yue2021prototypical} & 94.04 & 95.29 & 90.07 & 90.73 & 92.53  \\
    ~ & IS \cite{yue2021prototypical} & 85.07 & 92.42 & 82.50 & 83.82 & 85.95  \\
    ~ & SSL \cite{yue2021prototypical} & 93.67 & 95.14 & 88.82 & 90.22 & 91.96  \\
    \bottomrule
  \end{tabular}
  
\end{table*}

\item \textbf{T-SSL.} T-SSL consists of a feature encoder and a cosine similarity-based classifier. The output of the feature encoder is linearly embedded into 512-d and then $\ell_2$-normalized. The normalized vectors are used for $k$-means clustering and label prediction. We train models with learning rate $lr$ = 5e-5 and weight decay = 1e-7 for 10000 iters. The balancing coefficient $\lambda_{mim}$ is constantly assigned to 0.5 and $\lambda_{is}$ is adaptively updated from 0 to 0.1 using the scheme in \cite{ganin2015unsupervised}.

\item \textbf{BiT-DANN} BiT-DANN consists of a feature encoder, a label predictor, and a domain classifier. The domain classifier has the same architecture as the one in T-ADV. Similarly, we implement BiT-DANN on downstream tasks using ImageNet pre-trained BiT backbones. We train models with learning rate $lr$ = 3e-3 and weight decay = 5e-4 for 10000 iters except that the domain classifier use a learning rate $lr$ = 0.015.

\item \textbf{BiT-MME} BiT-MME consists of a feature encoder and a cosine similarity-based classifier. The classifier has the same architecture as the one in T-ADV. The learning rate of this part is also scaled up 10 times from the base set. The coefficient $\lambda_{e}$ is a constant value of 0.1. The other training setups are consistent with BiT-DANN.

\item \textbf{BiT-SSL} The architecture of BiT-SSL is consistent with T-SSL. except the backbone is replaced by a BiT model. We train models with learning rate $lr$ = 0.01 for 10000 iters. The balancing coefficient $\lambda_{mim}$ and $\lambda_{is}$ is respectively assigned to 0.5 and 1.0.

\end{itemize}

\begin{table}[t]
  \caption{\textbf{Results of Generalization-enhanced methods under texture shifts and style shifts.}}
  \label{table:additional DA results4}
  \centering
  \small
  \begin{tabular}{cccc}
    \toprule
    \multirow{2}{*}{Model} & \multirow{2}{*}{Method} &  Texture & Style \\ \cmidrule(lr){3-3} \cmidrule(lr){4-4}
    ~ & ~ & Stylized ImageNet & ImageNet-R  \\ \midrule
    \multirow{6}{*}{DeiT-S/16} & - & 13.11 & 25.91   \\
    ~ & T-ADV & \textbf{27.16} & \textbf{48.97}  \\
    ~ & T-MME & 17.90 & 47.71  \\ 
    ~ & T-MIM & 15.79 & 46.60  \\
    ~ & T-IS & 9.49 & 37.08  \\
    ~ & T-SSL & 13.78 & 47.33  \\ \midrule
    \multirow{6}{*}{BiT} & - & \textbf{6.05} & 23.46 \\
    ~ & DANN \cite{ganin2015unsupervised} & 4.49 & 35.33  \\
    ~ & MME \cite{saito2019semi} & 5.33 & 31.84  \\ 
    ~ & MIM \cite{yue2021prototypical} & 3.99 & 37.79   \\
    ~ & IS \cite{yue2021prototypical} & 3.46 & 28.36   \\
    ~ & SSL \cite{yue2021prototypical} & 4.15 & \textbf{38.29}   \\ 
    \bottomrule
  \end{tabular}
  
\end{table}

\subsubsection{T-SNE Visualization Implementation}

To begin with, we select five classes from DomainNet dataset for visualization: \emph{boomerang}, \emph{bowtie}, \emph{circle}, \emph{duck}, \emph{envelope}, since samples of these classes in \emph{quickdraw} are basically recognizable and ViTs could obtain decent performance. We then utilize samples of these classes in four domains to generate Class Token data of DeiT-S trained on \emph{real}. We extract Class Tokens of layer 8 and layer 12, and employ T-SNE to embed Class Tokens of four domains into the same feature space for each layer.

\section{More Experiment Results}

\subsection{Grad-CAM Visualization Results}

We conduct Grad-CAM visualization on more classes including \emph{bowtie}, \emph{coffee cup}, \emph{duck} and \emph{hourglass}, as shown in \cref{fig:more Grad-CAM}. All attention maps are generated using models trained on \emph{real}. We shall have the same observation as in the main text that DeiT-S is capable of capturing the key structural information of objects in whichever domain. Take for example the class \emph{hourglass}. DeiT-S consistently concentrate on the core area of a hourglass, \eg the glass portion. By contrast, BiT model tends to focus on the peripheral support in the case of \emph{real}, \emph{painting} and \emph{sketch}. The above results reconfirm the conclusion that ViTs focus more on structures.

\begin{figure}[t]
\vspace{-10pt}
\centering
\includegraphics[width=1\linewidth]{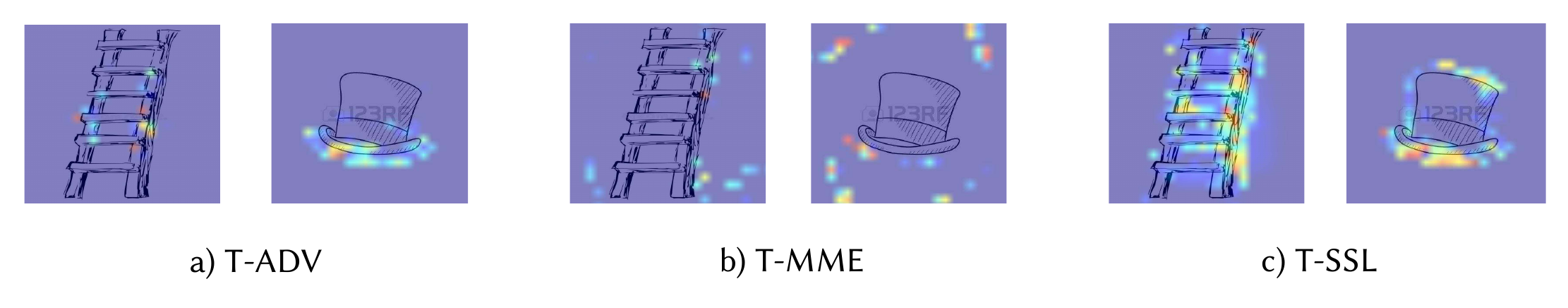}
\vspace{-15pt}
\caption{\textbf{Visualization of attention weights}}
\label{fig:att}
\vspace{-10pt}
\end{figure}

We further investigate the properties of the generalization-enhanced methods. Specifically, we generate the Grad-CAM attention maps of domain \emph{sketch} of T-ADV, T-MME and T-SSL, as illustrated in \cref{fig:att}. T-SSL can capture more comprehensive structure than two other methods, which may be the major factor why it outperforms T-ADV and T-MME.

\subsection{Ablation Study on Self-supervised Generalization Enhancing Method}
Since the self-supervised learning method T-SSL and BiT-SSL contains two major loss terms $\mathcal{L}_{\mathrm{IS}}$ and $\mathcal{L}_{\mathrm{MIM}}$ in enhancing out-of-distribution generalization, we separately test their effectiveness. As the results are shown in \cref{table:SSL ablation}, we could observe that 1) $\mathcal{L}_{\mathrm{MIM}}$ works the best for VGG-16, 2) but the combination of two parts perform the best on average for larger models including DeiT-S/16 and BiT. Thus, we could conclude that there exists a mutual promotion between the in-domain self-supervision and mutual information maximization towards large models.  

\subsection{Generalization-Enhanced ViTs Results under Multiple Shifts}
We examine the effectiveness of the generalization-enhanced methods under multiple shifts, including corruption shifts, texture shifts, and style shifts. We respectively use ImageNet-C, Stylized-ImageNet, and ImageNet-R for experiments. Because of the lack of training sets, we make a 2:1 split on these benchmarks and the ImageNet validation set for training and testing. Specifically, we use severity 3 of corruptions for use. The results are shown in \cref{table:additional DA results1}, \cref{table:additional DA results2}, \cref{table:additional DA results3} and \cref{table:additional DA results4}. From the results we could observe that 1) MME dominates the results under corruption shift for both types of models, 2) T-SSL performance the best under background shifts while DANN works the best for BiT models. 3) these generalization-enhancing methods may be harmful to generalization under certain distribution shifts for BiTs, e.g. defocus and brightness, while having little infection on DeiTs. 

\subsection{Relationship between the attention mechanism and the inductive bias}
\begin{figure}[h]
\centering
\subfloat[shape bias change]{
\includegraphics[width=0.45\linewidth]{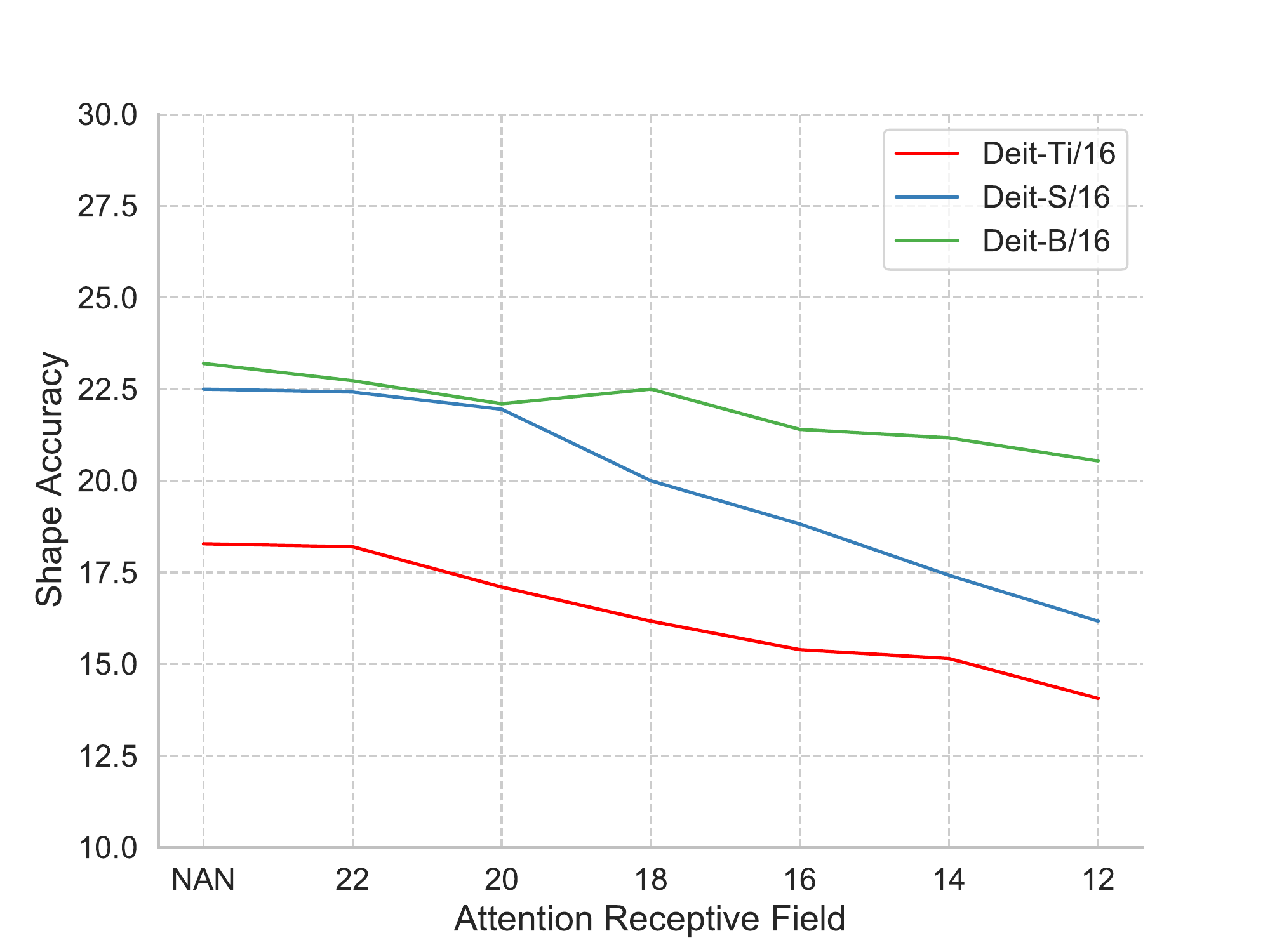}
}
\subfloat[texture bias change]{
\includegraphics[width=0.45\linewidth]{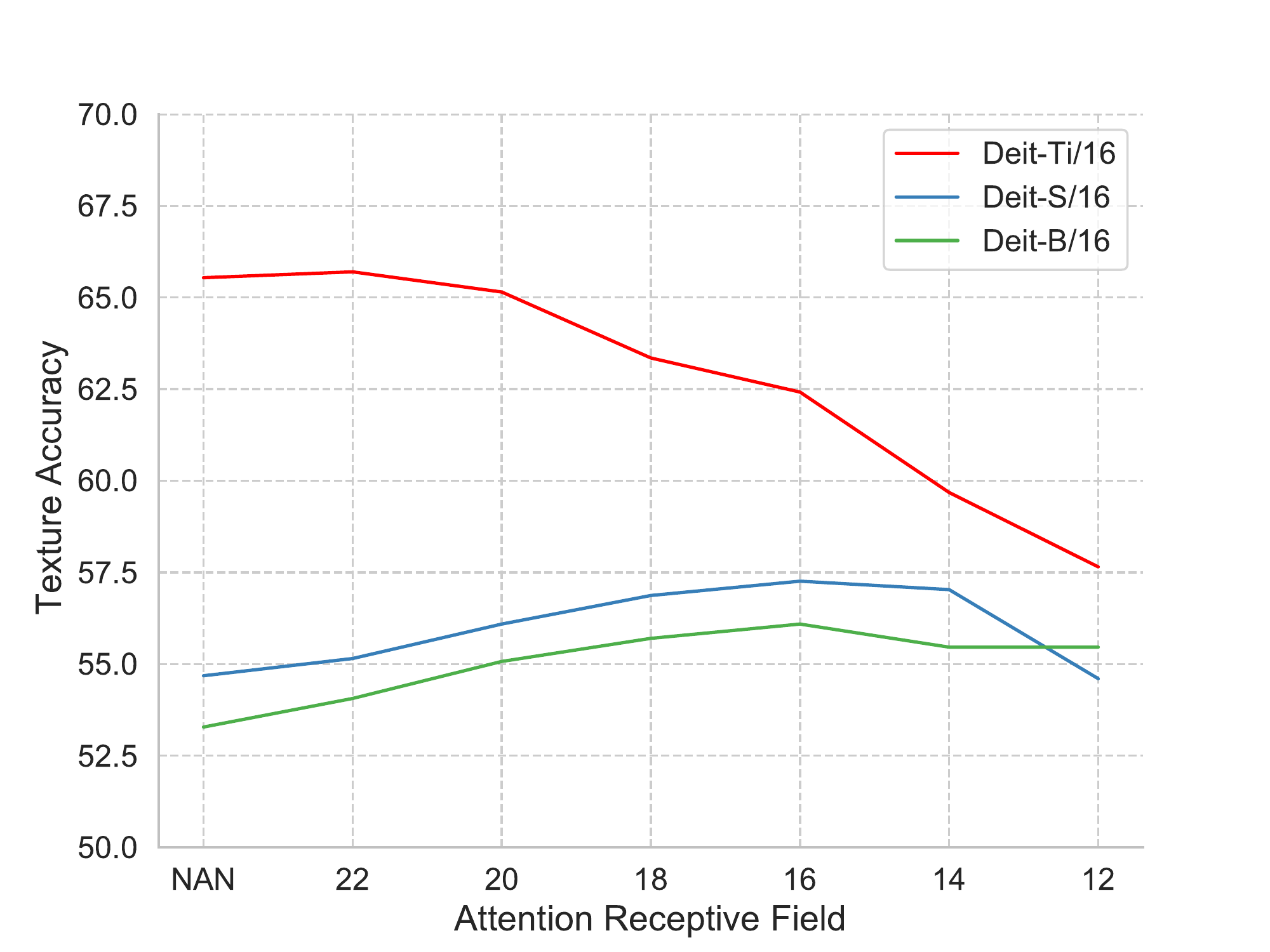}
}
\vspace{-8pt}
\caption{\textbf{Process of change of model shape and texture biases during the change of test-time attention receptive field sizes.}}
\label{fig:shape_text}
\vspace{-15pt}
\end{figure}

We further investigate how the attention mechanism, specifically the receptive field size of attention, contributes to the bias of ViTs. Due to the limited time, we make modifications on the well-trained ViTs using global attention by change the test-time attention receptive field size. As shown in \cref{fig:shape_text}, we observe the interesting phenomenon that model shape bias decreases as the attention receptive field size is reduced, while the model texture bias behave in the opposite way. For DeiT-S and DeiT-B, model texture bias even increases. Based on the above results, we could conclude that larger size of attention receptive field contributes to a better shape-biased ViT model.

\end{document}